%% file: main.tex
\pdfoutput=1

\documentclass[11pt]{article}

\usepackage{acl}

\usepackage{times}
\usepackage{latexsym}

\usepackage[T1]{fontenc}

\usepackage[utf8]{inputenc}

\usepackage{microtype}

\usepackage{makecell}
\usepackage{times}
\usepackage{latexsym}
\usepackage{tabularx}
\usepackage{array}
\usepackage{multirow}
\usepackage{siunitx}
\usepackage{booktabs}
\usepackage[pdftex]{graphicx}
\usepackage{enumitem}
\usepackage{xspace}
\usepackage{caption}

\usepackage{url}            
\usepackage{amsmath,amsthm,amsfonts,amssymb,bm,stmaryrd}       
\usepackage{nicefrac}       
\usepackage{enumitem}
\usepackage[noend]{algpseudocode}
\usepackage{algorithm}
\usepackage{mathtools}
\usepackage{nccmath}
\usepackage{multirow}
\usepackage{bigdelim}
\usepackage{color, colortbl}
\usepackage{xcolor}		
\definecolor{darkblue}{rgb}{0, 0, 0.5}
\hypersetup{colorlinks=true, citecolor=darkblue, linkcolor=darkblue, urlcolor=darkblue}
\usepackage{booktabs}
\usepackage{dcolumn}
\newcolumntype{d}{D{.}{.}{-1}}
\newcolumntype{z}{D{(}{\ (}{1.1}}
\usepackage{graphicx}
\usepackage{titlesec}
\usepackage{hyperref}
\usepackage{url}
\usepackage{framed}
\usepackage{tcolorbox}
\usepackage{microtype}
\usepackage{subcaption}
\usepackage{graphicx}
\usepackage[normalem]{ulem}
\useunder{\uline}{\ul}{}
\usepackage{fontawesome}
\usepackage{calc}

\title{
Stretching Sentence-pair NLI Models\\ to Reason over Long Documents and Clusters}

\author{Tal Schuster,$^1$ Sihao Chen,$^{1,2}$ Senaka Buthpitiya,$^1$ Alex Fabrikant,$^1$ Donald Metzler$^1$\\
\\
    $^1$Google Research \qquad $^2$University of Pennsylvania \\
    
    {\tt \{talschuster,sihaoc,senaka,fabrikant,metzler\}@google.com}}

\input{macros}
\begin{document}
\maketitle
\begin{abstract}
Natural Language Inference (NLI) has been extensively studied by the NLP community as a framework for estimating the semantic relation between sentence pairs. While early work identified certain biases in NLI models, recent advancements in modeling and datasets demonstrated promising performance.
In this work, we further explore the direct zero-shot applicability of NLI models to real applications, beyond the sentence-pair setting they were trained on. First, we analyze the robustness of these models to longer and out-of-domain inputs. Then, we develop new aggregation methods to allow operating over full documents, reaching state-of-the-art performance on the ContractNLI dataset. Interestingly, we find NLI scores to provide strong retrieval signals, leading to more relevant evidence extractions compared to common similarity-based methods. Finally, we go further and investigate whole document clusters to identify both discrepancies and consensus among sources. In a test case, we find real inconsistencies between Wikipedia pages in different languages about the same topic.\footnote{Released Wikipedia translated clusters dataset: \url{https://github.com/google-research-datasets/wiki-translated-clusters-nli}}
\end{abstract}

\input{sections/intro}

\input{sections/model}

\input{sections/beyond_sent}

\input{sections/eval_datasets}
\input{sections/experiments}

\input{sections/related}
\input{sections/conclusion}

\section*{Acknowledgements}
We thank Shashi Narayan and Simon Baumgartner for valuable feedback on the writing. We also thank Sumit Sanghai, Annie Louis, Jiaming Luo, Roee Aharoni, Yi Tay, Kai Hui, Jai Gupta, Vinh Tran, and Dara Bahri for helpful conversations and feedback.

\section*{Limitations}
As mentioned in the conclusion section and along the paper, our experiments focus on exploring and leveraging the direct signal from sentence-pair NLI models. In this work, we did not employ more advanced techniques to process the data such as contextualizing the premise or fragmenting the hypothesis. We leave such studies on further improving the downstream performance to future work. Also, we train and evaluate our NLI models on English inputs, and don't explore morphologically richer languages here.

In our Wikipedia clusters experiments, we translate all pages to English. This translation process might introduce some mistakes. However, when examining several samples we find the translation quality to be overall high. Also, capturing translation mistakes with \ourmodel is another potentially interesting application of our setup. 

Finally, our zero-shot evaluations on DocNLI and ContractNLI are out-of-domain, but the Wikipedia Cluster experiments are partly in-domain as the training data includes premises from Wikipedia. Yet, different from the training, the hypothesis is also a Wikipedia sentence. Also, we use the VitaminC test set to avoid potential overlaps with pages that the model saw on training.

\section*{Ethical Considerations}
We emphasize that our method and experimentation on identifying disagreements between documents is focused on the research question of whether our models can capture the required signal from the text. When highlighting such cases, we do not claim by any means to state anything regarding the truthfulness of any of the statements. Rather, we examine the question regarding the usefulness of ranking the statements by their perceived agreement with each other.

\bibliography{anthology,custom}

\clearpage
\appendix
\counterwithin{figure}{section}
\counterwithin{table}{section}
\input{sections/app_technical}
\input{sections/app_binary}
\input{sections/app_retrieval}
\input{sections/app_dataset}
\input{sections/app_wiki}


\end{document}

%% file: macros.tex
\newcommand{\STAB}[1]{\begin{tabular}{@{}c@{}}#1\end{tabular}} 

\floatname{algorithm}{Algorithm}

\DeclareMathOperator*{\argmax}{arg\,max}

\DeclareMathOperator{\softmax}{Softmax}

\newcommand{\newpar}[1]{\noindent\textbf{{#1}~}}

\newcommand{\ourmodel}{\textsc{seNtLI}\xspace}
\newcommand{\prem}{X_{\mathrm{prem}}}
\newcommand{\hyp}{X_{\mathrm{hyp}}}
\newcommand{\func}{f}
\newcommand{\summac}{\textsc{SummaC}\xspace}

\newcommand{\parhead}[1]{\noindent\textbf{#1}}


%% file: sections/intro.tex
\section{Introduction}\label{sec:intro}

Natural Language Inference (NLI) involves automatically determining whether the meaning of one piece of text (i.e., hypothesis) can be inferred from another (i.e., the premise) \citep{dagan2006}. 
This formulation is relatively simple, enabling large-scale data annotation~\citep[e.g.,][]{bowman-etal-2015-large, williams-etal-2018-broad}, yet imposes complex semantic reasoning challenges (e.g., background knowledge, commonsense), leading to a useful training and evaluation NLP framework~\citep{mishra-etal-2021-looking, Sainz2022textual, vu-etal-2021-strata}.





Formally, in NLI, we are interested in learning a function $\func: (\hyp \times \prem) \rightarrow \mathcal{Y}$ that predicts the relation $Y$ between the provided premise $\prem$ and the examined hypothesis $\hyp$, where $\mathcal{Y} = \{\mathrm{entailment, neutral, contradiction}\}$.\footnote{This formulation also fits the task of fact-checking a claim against given evidence~\citep{thorne-etal-2018-fever}, therefore henceforth we use the NLI terminology for both tasks.}
In most NLI datasets, $\hyp$ and $\prem$ are short texts consisting of one or few sentences, allowing current Large Language Models (LLMs) with limited input length to process the two with cross-attention. In practice, however, many systems require operating over long texts such as full documents or even collections of documents without knowing a priori which parts are most relevant.

\begin{figure*}[t]
    \centering
    \includegraphics[width=0.95\textwidth]{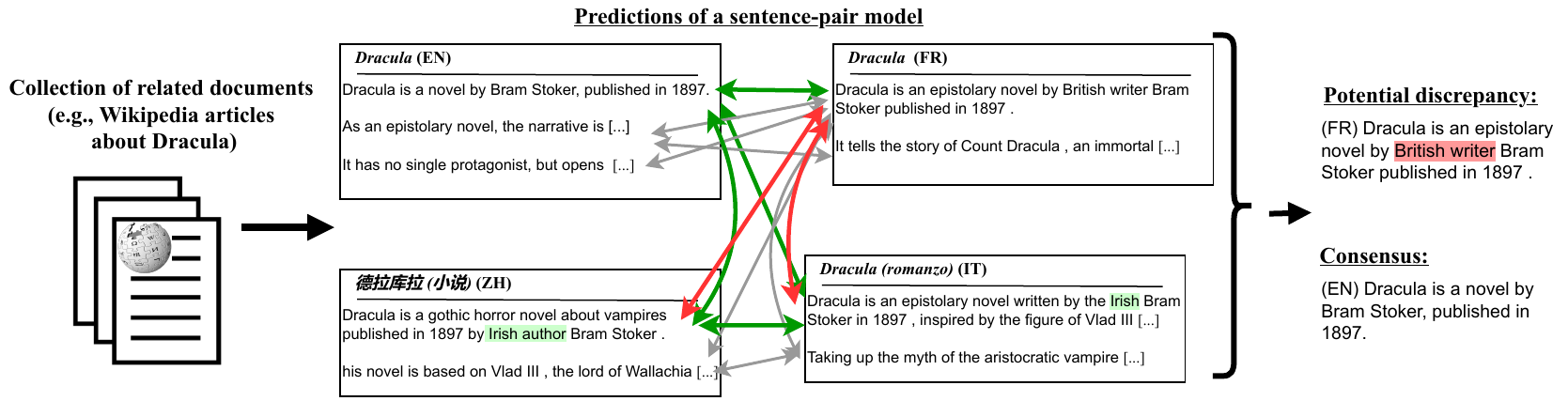}
                   \vspace{-5pt}
    \caption[]{Illustration of our procedure for flagging potential discrepancies in document clusters (\S\ref{sec:clusters_method}). This is a real-world example from Wikipedia's translated articles in different languages about the novel Dracula (as of Feb.\ 2022).\footnotemark\ Our model identified the French Wikipedia to disagree with other articles on the nationality of the author.}
    \label{fig:intro}
                       \vspace{-8pt}
\end{figure*}

Consider the example in Figure~\ref{fig:intro}. The system is trying to reason over a collection of documents and find statements that they all agree upon (consesnus), or alternatively, find potential disagreements across documents. Instead of having a clear hypothesis-premise pair, each statement across all documents is an hypothesis of interest that should be evaluated against all other documents as the premise.

In this paper, we focus on these realistic scenarios and present retrieve-and-classify methods for inferring over long and out-of-distribution (OOD) inputs in a zero-shot fashion. As this setting emphasizes the need for a robust sentence-pair NLI model as a backbone, we train on multiple datasets, including adversarial and contrastive ones~\citep{nie-etal-2020-adversarial,schuster-etal-2021-get} to increase the model's robustness and avoid dataset-specific biases~\citep{gururangan-etal-2018-annotation,mccoy-etal-2019-right,poliak-etal-2018-hypothesis, schuster-etal-2019-towards}.



Our proposed pipelines go beyond the length and format supported by most LMs and include new retrieval, aggregation and classification solutions---all based on the same classifier. Thereby, we continue the line of work on evaluating the robustness of such models and their ability to truly capture semantic relations. Moreover, long inputs highlight the commonly overlooked yet practically important challenge of specifying ``neutral'' relations. For example, mistaking certain part of the premise as entailing instead of neutral can overshadow another segment that contradicts the hypothesis.

We first evaluate the zero-shot performance of our multi-task NLI model on $\prem$s that are longer than the examples seen in training, but short enough to allow supervised LMs with similar length constraints to perform well~\citep{yin-etal-2021-docnli}. In the zero-shot setting, we find that models generalize beyond the training distribution, but drop in performance for very long $\prem$s ($>400$ tokens).


\footnotetext{\url{https://fr.wikipedia.org/w/index.php?title=Dracula&oldid=190970820}}

Then, we turn to focus on the scenario of having a full document as $\prem$~\citep{koreeda-manning-2021-contractnli-dataset}. A typical approach---the default behavior of most LMs---would truncate the end of the document beyond some predefined length (e.g., 512 tokens). However, this might remove and ignore important information.
Instead, we hypothesize that a good NLI model should be able to separate the wheat from the chaff and distinguish neutral spans towards $\hyp$ from informative ones. To this end, we develop a solely NLI-based retrieve-and-classify approach that outperforms similarity-based retrievers and whole-document classifiers.

Finally, we go further and demonstrate the utility of our model for reasoning over entire clusters of related documents. Our proposed procedure, illustrated in Figure~\ref{fig:intro}, ranks all of the cluster's spans by their entailment relations with spans from other documents. Testing our approach on Wikipedia introductions on the same topic in different languages, we successfully identify claims that are unique to one version and contradicted by others.

In summary, this work stretches sentence-pair NLI models to new practical capabilities and demonstrates their direct utility in real-world applications.  
Our main contributions include: \vspace{-0.5em}
\begin{itemize}[leftmargin=*]
    \item A multi-task sentence-level NLI model with strong zero-shot and supervised performance for both evidence retrieval and classification. \vspace{-0.5em}
    \item Simple and effective retrieve-and-classify methods to extend sentence-pair semantic classifiers and outperform  whole-document models.\vspace{-0.5em}
    \item A new entailment task and dataset that requires inference over clusters of documents (\S\ref{sec:wiki_clusters}).\footnote{We intend to release this new dataset upon publication.} \vspace{-0.5em}
    \item Demonstrating the utility of our approach to reveal real and simulated discrepancies in Wikipedia pages by automatically comparing with content from multiple translated articles.
\end{itemize}

%% file: sections/model.tex
\section{Model and Definitions}\label{sec:model}

Our pipelines build on entailment scores for hypothesis-premise pairs.
To predict these scores, we train a sentence-pair NLI model that we use as the backbone for all methods. Specifically, we pick the T5 encoder-decoder architecture as it has been shown to perform well in multi-task and transfer settings~\citep{arib2021ext5, t5_paper}.
See Appendix~\ref{sec:technical_det} for more technical details.

\newpar{Definitions.}
As T5 is a seq-to-seq model, we train it over the training set $(\hyp, \prem, Y) \sim \mathcal{D}_{\mathrm{train}}$ by feeding the following format to the encoder:
``\emph{entailment: $\hyp$ [SEP] $\prem$}.'' The decoder's goal is to generate a single character 'e', 'n', or 'c', representing 
the three classes in $\mathcal{Y}$: entailment, neutral, and contradiction, respectively. Thereafter, when making a prediction on a new pair $(\hyp, \prem) \in \mathcal{D}_{\mathrm{test}}$, we encode the input and measure the decoder's score $s_y$ for each of the three classes. Finally, we normalize the three scores with a softmax operator: $p_y=\softmax(s_e, s_n, s_c)[y]$. Note that  these scores should not be directly treated as class probabilities as they are not calibrated, especially when evaluating out-of-domain inputs. However, as we will show in our experiments, they can be readily tuned or leveraged as a valuable signal for textual semantic relations.


\newpar{The \ourmodel model.}
To train our model, we use the following sentence-pair datasets: 
 SNLI~\citep{bowman-etal-2015-large}, MNLI~\citep{williams-etal-2018-broad}, ANLI~\citep{nie-etal-2020-adversarial}, FEVER~\citep{thorne-etal-2018-fever}, and VitaminC~\citep{schuster-etal-2021-get}. For FEVER, we use the sentence-pair version from VitaminC with retrieved evidence to neutral claims.

This multi-task training is important for improving the robustness of the model by leveraging a large amount of diverse data. Also, including adversarial (ANLI) and contrastive (VitaminC) examples was shown to prevent the model from relying on hypothesis-only biases. Table~\ref{tab:train_datasets} shows the input length statistics of the datasets according to the English sentence-piece tokenizer of T5. 
Having concise and short inputs generally makes the task less ambiguous---as claim in question is clear---and focused on the textual entailment component.

We denote our T5 mutli-task NLI model trained on all 5 datasets \ourmodel. Our application-specific pipelines will differ in the way that they make use of \ourmodel's predictions. 
We also experiment with models trained only on MNLI without multi-tasking (-M.T), only on the three *NLI datasets (-F.V), or with ContractNLI~\citep{koreeda-manning-2021-contractnli-dataset} examples.




\begin{table}[t]
\small
    \centering
    \resizebox{\linewidth}{!}{
    \begin{tabular}{c|zzc}
\toprule
        Train dataset & \multicolumn{1}{c}{Hypothesis length} & \multicolumn{1}{c}{Premise length} & \multicolumn{1}{c}{Train pairs}   \\
        \midrule
        MNLI & 13.23 (7-20) & 27.57 (9-50) & 392,702 \\
        SNLI & 9.50 (5-15) & 16.86 (9-27) & 550,152\\
        ANLI & 13.32 (7-21) & 79.95 (53-111) & 162,865\\
        FEVER & 12.50 (8-18) & 47.98 (21-82) & 178,059\\
        VitaminC & 18.18 (10-29) & 43.03 (19-72) & 370,653 \\
    \bottomrule
    \end{tabular}
        }%
                       \vspace{-5pt}
    \caption{Training datasets of \ourmodel. We report the average length of the tokenized (for T5) hypothesis and premise, and the 10th-90th percentiles in parentheses.}
    \label{tab:train_datasets}
      \vspace{-10pt}
\end{table}

%% file: sections/beyond_sent.tex
\section{Beyond Sentence-level Inference}\label{sec:beyond_sent}


We now assume that our target application requires the evaluation of hypotheses against \emph{long texts}. 
We also assume that we have limited or no training data for this domain, and focus on \emph{zero-shot} transfer.
Formally, we assume that $\prem$ is a document consisting of $n$ sentences $S_{1:n}$ and we don't know which part of the document is most relevant for verifying or rejecting $\hyp$. 
In this case, it is common to not only classify the truthfulness of the given statement, but to also point to the exact evidence in the document that led to this conclusion. 
This is a crucial requirement, both benefiting the interpretability and trustworthiness of the model (e.g., avoiding hypothesis-only bias), and saving human time needed for manual prediction verification.

\subsection{Na\"ive premise truncation}
A na\"ive design choice for such applications would be to simply use a similar cross-attention model and provide as much as possible from the input text. As Transformer models are trained with a defined maximum input length limit (typically 512 tokens), this approach has obvious limitations. Yet, in some applications we can assume where the relevant information is likely to be and remove the rest. For example, \citet{yin-etal-2021-docnli} found a model that truncates the input to even outperform a model that supports long inputs on DocNLI. 

Nevertheless, this approach is unlikely to suit very long inputs as the complexity of Transformers grows quadratically with the input length~\citep{tay2021long, attention_all_you_need}. Also, this approach doesn't directly support the important interpretability requirement discussed above, as it is unclear which part of the long document led the model to its prediction.






\subsection{Retrieve-and-classify over long premises}
\label{sec:ret-and-cls}

Instead, we opt to break the long premise into individual sentences and make pointwise predictions against the hypothesis before aggregating them to the final classification. This approach can readily extend to any document length without modifications to the core NLI model. Also, in zero-shot transfer, this allows better alignment with the sentence-pair training distribution. While it requires $n$ inference runs of the NLI model instead of a single pass, the cost increases linearly with $n$, unlike the quadratic effect of increasing the input length. Also, these inference passes can be computed in parallel with batches.

This pointwise approach, however, has some limitations. First, it requires a robust sentence-pair model that can separate neutral sentences from relevant ones. Second, it doesn't immediately support multi-hop inference over multiple sentences~\citep{jiang-etal-2020-hover}.  This can be partially alleviated with preprocessing techniques \cite[e.g.,][]{choi-etal-2021-decontextualization}. In practice, we don't observe this limitation in our explored applications as most sentences in these domains are sufficiently self-contained (see Figure~\ref{fig:intro}).

Next, we discuss methods for performing the two key steps of the retrieve-and-classify approach.

\subsubsection{Retrieval}\label{sec:retrieval}
Given the long multi-sentenced premise, we would like to identify which sentences are most helpful for accepting or rejecting the hypothesis. We focus on methods supporting data-sparse target domains. 

\newpar{Similarity-based retrieval.}
Most unsupervised retrievers use a similarity function (e.g., inner product) over sparse or dense representations to compare $\hyp$ and each of the premise sentences.

\newpar{NLI-based retrieval.}
We introduce an alternative retrieval approach where we use the non-neutral scores of an NLI model to determine the usefulness of a sentence for classifying the hypothesis. This is motivated by the fact that two neutral sentences can be very similar by many measures, but uninformative for our purpose, such as, e.g., $\prem=$``$\textit{We discuss later whether } \hyp$''. 

Here, each sentence ends up having two ranking scores with respect to the hypothesis: for entailment and for contradiction. As we see next, this granularity is useful for downstream steps such as reranking or reasoning over clusters.


\subsubsection{Classification}
Following the evidence retrieval, we define two methods for constructing the final prediction:


\newpar{Retrieve-and-Predict.}
Assuming NLI-based retrieval, we can simply reuse the same scores.
We pick the span with the strongest score for each label, $\bar{p_y} = \max_{i \in [1,n]}{(p_{y,i})}$, and then predict by the highest NLI score: 
\begin{equation*}
\begin{small}
\text{prediction} =
\left\{
	\begin{array}{ll}
		\argmax_y \bar{p_y}  & \mbox{if } \max_{y\in \{e, c\}} \bar{p_y} > T,  \\
		\text{'neutral'} & \mbox{otherwise.}
	\end{array}
\right.
\end{small}
\end{equation*}
We simply set $T=0.5$ for the zero-shot setting, but it can be potentially tuned. Effectively, the prediction is determined by the span with the strongest sentiment towards the hypothesis.

\begin{figure}[t]
    \centering
    \includegraphics[width=0.48\textwidth]{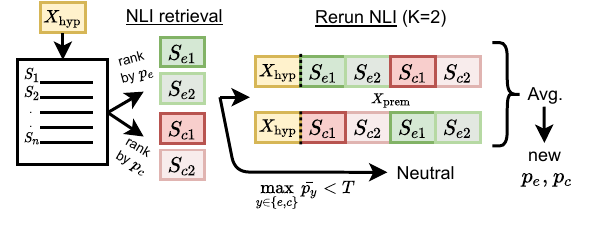}
      \vspace{-16pt}
    \caption[]{NLI-based retrieve-and-rerank concatenates the K spans with the strongest 'entail' score, and the K with the strongest 'contradict' score for reranking, as long as some span's non-neutral score exceeds $T$.}
      \vspace{-8pt}
    \label{fig:rerank}
\end{figure}

\newpar{Retrieve-and-Rerank.}
Instead of directly using the retrieval scores, we rerun the same NLI model on the original hypothesis, and a concatenation of the top-K spans retrieved for both non-neutral labels. For symmetry, we average over two instances: the first concatenating the top-entailing spans in score order, then the top-contradicting spans; and the second instance switching the entailing and contradicting spans. Figure~\ref{fig:rerank} illustrates this process with $K=2$.

This reranking allows the NLI model to directly contrast the spans that are most entailing with the spans that are most contradicting toward the hypothesis.
The resultant multi-sentence premise is longer than the training distribution. Yet, we find \ourmodel to generalize well to slightly longer but focused premises.


\subsection{Multi-sentence hypotheses}
\label{sec:multi_sent_hyp}

NLI models could also be useful in scoring texts that are longer than a single focused statement. Recently, \citet{SummaC} used NLI scores to predict the faithfulness of generated summaries with respect to their source. Here, we can break both the premise (e.g., the source) and the hypothesis (the summary) into spans, retrieve-and-classify (with or without reranking) for each hypothesis span, and aggregate.
We consider two methods for aggregating the scores of the hypothesis spans:

\newpar{Soft aggregation.}
Following \citet{SummaC}, we take the average entailment scores across spans. When using our reranking method, we take the shifted difference between the scores: $p_e - p_c + T$.

\newpar{Hard aggregation.}
We take the minimum entailment score across spans, effectively requiring all of the hypothesis spans to be strongly supported.

\subsection{Reasoning over multi-document clusters}\label{sec:clusters_method}

\begin{algorithm}[t]
\small
\caption{Find factual discrepancies in a cluster.}\label{alg:cluster}
\begin{algorithmic}
\Require Cluster of documents $\mathcal{D}$ with spans $\mathcal{S}$, and NLI model.
\Ensure Sorted spans by discrepancy likelihood.
\State $\omega_{i,j} \gets 0 \quad \forall \text{ span } j \text{ in document } i$
\For{$D_i \in \mathcal{D}$}
\For{$S_j \in D_i$}
\State $\hyp \gets S_{i,j}$; $\quad \Omega \gets \{ \}$ 
\For{$D_k \in \mathcal{D} \setminus D_i$}
\State $\Gamma \gets \{ \}$ 
\For{$S_{k,l} \in D_k$}
\State $p_c \gets \text{\ourmodel}(\hyp, S_{k,l})[c]$
\State $\Gamma \gets \Gamma \cup \{p_c\}$
\EndFor
\State $ \Omega \gets \Omega \cup \max(\Gamma)$ 
\EndFor
\State $\omega_{i,j} \gets \mathrm{mean}(\Omega)$ 
\EndFor
\EndFor
\Return $\mathcal{S}$ sorted by respective $\omega$
\end{algorithmic}
\end{algorithm}

\begin{table*}[t]
\small
    \centering
    \resizebox{0.95\linewidth}{!}{
    \begin{tabular}{c|c|cccc}
        \toprule
        Eval dataset & \multicolumn{0}{c}{Hypothesis length} & \multicolumn{0}{|c}{Premise length} & Docs per premise & \multicolumn{0}{c}{Sents per doc} & \multicolumn{0}{c}{Sentence length}\\
        \midrule
        DocNLI (-ANLI) \citep{yin-etal-2021-docnli} & 98.91 (52-144) & 530.24 (70-1312) &  1 & 17.53 (3 - 43)  & 30.47 (12-51)
 \\ 
        ContractNLI  \citep{koreeda-manning-2021-contractnli-dataset} & 19.35 (11-27) & 2408.75 (939-4409) & 1 & 79.63 (36 - 128) & 30.28 (1-74)
\\ 
        \summac  \citep{SummaC} & 62.63 (22-117) & 678.52 (240-1234) & 1 & 22.16 (8-41) & 30.38 (11-50)\\ 
        Wiki Clusters (Section~\ref{sec:wiki_clusters}) & 34.83 (17-58) & 2498.39 (999-4152) & 9.86 (10-10) & 7.79 (2-16) & 32.52 (13-56)\\
    \bottomrule
    \end{tabular}
        }%
                \vspace{-5pt}
    \caption{Evaluation datasets and the average tokenized lengths (with 10/90th percentiles), number of sentences per premise document, and length of each sentence. The premise in our Wiki clusters consists of multiple documents.}
    \label{tab:eval_datasets}
      \vspace{-10pt}
\end{table*}

So far, we dealt with evaluating a single, short or long hypothesis against a long premise. In some applications, however, the user might not know which hypothesis to check, but rather would like to query over their corpus to identify the most extreme ones. 
For example, consider a collection of news articles on the same topic written by different sources.
Typical questions to ask about this corpus could be: \emph{``is there any claim made by one article that other articles disagree with?''}, or other queries like \emph{``what is the most controversial claim?''} or \emph{``is there consensus on some claims in the corpus?''}

Answering such questions goes beyond the typical NLI setting and requires understanding a complex many-to-many relation between the documents, involving multiple alignment and reasoning challenges. Therefore, any solution with low signal-to-noise ratio is likely to fail.

Using our robust \ourmodel model, we introduce an algorithm for identifying such claims. 
Algorithm~\ref{alg:cluster} ranks all of the cluster's spans by discrepancy likelihood.\footnote{When looking for consensus, $p_c$ is replaced by $p_e$.}
Each span is compared against all other spans from all documents. The score is determined by the most contradicting pairing from each document and averaged across the cluster. While this procedure requires many calls to the NLI model (quadratic in number and size of documents), they are independent and can easily be batched and parallelized. 


%% file: sections/eval_datasets.tex
\section{Evaluation Tasks and Datasets}\label{sec:eval_tasks}
We evaluate our methods on the following 3 benchmarks that contain 9 datasets from different domains. In addition, we create a new of its kind dataset with clusters of related documents (\S\ref{sec:wiki_clusters}).
The statistics are summarized in Table~\ref{tab:eval_datasets}.

\noindent\textbf{DocNLI}~\citep{yin-etal-2021-docnli} includes long hypotheses and premises, mostly from the news domain. We remove ANLI since it was included in our training data.\footnote{The ANLI examples cover only 1.2\% of the DocNLI test set, so the difference is minimal. \ourmodel and \ourmodel$_{\text{tuned}}$ get .350 and .410 $F_1$ scores on the full test set, respectively.}
Despite the length, a model with input limit of 512 tokens can generally perform well.


DocNLI uses only two classes, ``entail'' vs.\ ``not entail''.
We discuss different zero-shot conversion techniques from 3-way models to binary classification in Appendix~\ref{sec:binary_aggr}.

\noindent\textbf{ContractNLI}~\citep{koreeda-manning-2021-contractnli-dataset} has NLI examples in the legal domain. Each hypothesis is short and focused, but the premise is a long document (80 sentences on average). A model with input limit of 512 tokens performs poorly here.


\noindent\textbf{\summac}~\citep{SummaC} is a benchmark for predicting the factual consistency of summaries with their source.
We follow the zero-shot setting here, but for fair comparison with \citet{SummaC}, we also tune the threshold on the validation set for each dataset, and report the results on the test set. In early exploration we found naive threshold settings to be competitive as well.

\subsection{Wikipedia clusters evaluation dataset}\label{sec:wiki_clusters}

In addition, we create a new dataset for exploring \emph{inference over collections of related articles}.
Specifically, we collect clusters of introductions to popular Wikipedia articles on the same topic written in up to 11 different languages, machine translated to English. 
See App.~\ref{sec:wikiclusters_details} for more details.\footnote{The Wikipedia clusters data is available at: \url{https://github.com/google-research-datasets/wiki-translated-clusters-nli}}

Each version of Wikipedia is managed by a different community, leading to occasional disagreements or mistakes~\citep{iv2021fruit, unified_wiki}, or even the risk of version-specific conspiracy theories.\footnote{\url{www.bbc.com/news/technology-59325128}}
Therefore, automatically comparing and contrasting the information from different articles could be very helpful.
We examine both synthetic corruptions and real discrepancies.



\newpar{Corrupted articles.}
We simulate a corruption to the English version of each article by inserting a local edit to one of the sentences. The task is to use the other articles of that cluster to identify the sentence that was changed. While it is possible that all other articles don't mention any information about the specific corrupted fact, thanks to the popularity of the chosen articles and languages we find that mostly at least one of the articles includes sufficient information to refute the corrupted sentence. 

To create the corruptions, we use edits from the test set of the VitaminC dataset~\citep{schuster-etal-2021-get} that express opposite relations towards a mutual claim. 
In total, we create 824 instances based on 144 different topics. In each instance, we corrupt a single sentence from one of the English articles, and provide the 10 related articles from other languages to help identify which fact was changed.

We note that \ourmodel observed Wikipedia sentences (from other articles) in the training mixture. However, they were only used as the premise, whereas here they also represent the hypothesis.

\newpar{Real discrepancies.}
We look for discrepancies in-the-wild, searching in current Wikipedia. 
Here, we don't know which article, if any, might include a discrepancy. Therefore, in this setting, we focus on qualitative evaluation and explore whether we can rank all spans from all articles to identify real discrepancies, or consensus.




%% file: sections/experiments.tex
\begin{table}[t]
\small
    \centering
    \begin{tabular}{l|cc}
    \toprule
    Model & Dev. & Test \\
    \midrule
    Random & .198 & .199 \\
    \midrule
    \multicolumn{3}{l}{\underline{\textit{supervised:}}} \\
    Longformer-base$^*$~\citep{yin-etal-2021-docnli} & .462 & .444  \\
     Roberta-large~\citep{yin-etal-2021-docnli} & .631 & .613  \\
     T5-large & .642 &.618 \\
     \midrule
    \multicolumn{3}{l}{\underline{\textit{zero-shot:}}} \\
     \ourmodel (no setnence split) & .341 & .345 \\     
     \ourmodel$_{\text{tuned}}$  (no setnence split) & .409  & .408 \\  
     \bottomrule
    \end{tabular}
                   \vspace{-5pt}
    \caption{$F_1$(E) scores on the \textsc{DocNLI}(-ANLI) binary classification dataset. The zero-shot shot predictions are based on a threshold $T$ on the entailment score which is either set to $0.5$ or \emph{tuned} over $0.2\%$ of the dev set. $^*$Longformer's scores are over the full DocNLI.}
    \label{tab:res_docnli}
    \vspace{-8pt}
\end{table}

\section{Experiments}\label{sec:experiments}

We evaluate our inference pipelines against supervised models from DocNLI and ContractNLI, and the zero-shot \summac model, adopting the main evaluation metrics from each paper. We also train T5-large supervised models to directly compare with \ourmodel's zero-shot performance.

DocNLI~\citep{yin-etal-2021-docnli} used a RoBERTa-large~\citep{roberta} model with input limit of 512 tokens and a Longformer-base~\citep{longformer} model.
ContractNLI introduced the SpanNLI~\citep{koreeda-manning-2021-contractnli-dataset} model to process long documents that they train to jointly identify key spans and to make the final verdict.
\summac~\cite{SummaC} used a BERT-large~\citep{devlin-etal-2019-bert} model that was also trained on multiple NLI datasets.


\subsection{Zero-shot transfer to new domains}


\parhead{DocNLI.} 
Table~\ref{tab:res_docnli} summarizes the results on DocNLI.  
\ourmodel performs much better than a random baseline and is competitive with some supervised models, indicating promising transfer potential.


\begin{figure}[t]
    \centering
    \includegraphics[width=0.46\textwidth]{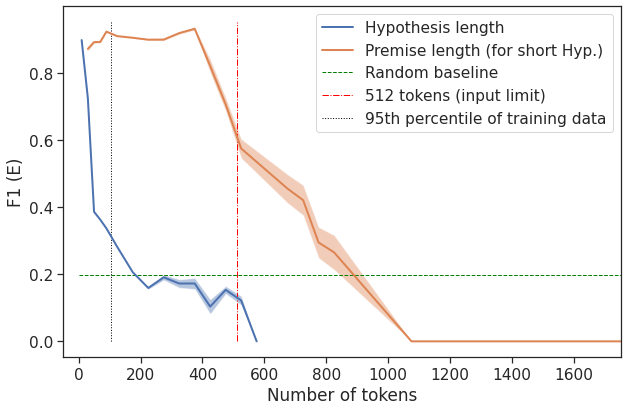}
        \vspace{-4pt}
    \caption[]{Effect of input length of DocNLI examples when naively using zero-shot \ourmodel \textbf{without sentence splitting} (simply providing the whole document as a single premise). The blue line shows $F_1$ score as a function of hypothesis length ($|\hyp|$). The orange line bins examples by the premise length ($|\prem|$), focusing only on short hypotheses (where $|\hyp|\le20$). For short hypotheses, the performance stays high even beyond the training distribution, but sharply drops around the input length limit of the model.}
    \label{fig:docnli_by_len}
     \vspace{-5pt}
\end{figure}

We also examine the effect of the input length on the performance in Figure~\ref{fig:docnli_by_len}.
First, we see that the performance is highly affected by the length of the hypothesis. \citet{yin-etal-2021-docnli} observed a similar trend even with supervised models. We conjecture that this is due to the natural increase in ambiguity with the hypothesis' length, as it is more likely to include multiple claims that could be questioned. 

Second, we look at the performance as a function of the premise length. To focus on examples where the hypothesis is well defined, we only consider cases with a short hypothesis of no more than 20 tokens (total of 5,932 cases from DocNLI(-ANLI) test set). As the orange line shows, \ourmodel performs well on these cases even when the premise is much longer than the inputs that the model was trained on, demonstrating promising potential for zero-shot applications. However, the performance significantly drops when reaching the input length limit, requiring us to truncate the premise.


\begin{table}[t]
\small
    \centering
    \begin{tabular}{ll|c|ccc}
    \toprule
    & Model & S.P. & $F_1$(C) & $F_1$(E) & AVG \\
    \midrule
    & Majority vote$^{\ddag}$ &  &.239 & .645 & .442  \\
    \midrule
    \multirow{4}{*}{\STAB{\rotatebox[origin=c]{90}{\textit{supervised}}}} 
        & SpanNLI-base$^{\ddag}$ & & .657 & .816 & .736 \\
        & SpanNLI-large$^{\ddag}$ & &.620 & .806 & .713 \\
        & T5-large & v &\textbf{.815} & .971 & .893\\ 
        & \ourmodel$_{\text{+ContractNLI}}$ & v & .813 & \textbf{.978} & \textbf{.895}\\
     \midrule
    \multirow{3}{*}{\STAB{\rotatebox[origin=c]{90}{\textit{0-shot}}}} 
        &\ourmodel$_{\text{-M.T}}$ & v &.616 & .882 & .749 \\
        & \ourmodel$_{\text{-F.V}}$ & v &.616 & .869& .742 \\
        & \ourmodel & v & \textbf{.661} &\textbf{.904} & \textbf{.782} \\ 
     \bottomrule
    \end{tabular}
    \vspace{-5pt}
    \caption{ContractNLI results with Oracle evidence spans (excluding neutral examples). Sentence-pair models (S.P.), even in zero-shot setting, outperform the SpanNLI model that was trained on long inputs. $^{\ddag}$Results from \citet{koreeda-manning-2021-contractnli-dataset}.}
    \label{tab:contnli_oracle}
    \vspace{-4pt}
\end{table}

\parhead{ContractNLI.} 
We test zero-shot transfer to the legal domain of ContractNLI. To disentangle the effect of input length, we first examine an oracle retriever setting where the premise includes only the few relevant sentences. Table~\ref{tab:contnli_oracle} summarizes the results.
Surprisingly, we find our zero-shot sentence-pair models outperform the supervised SpanNLI models. \ourmodel, trained on all five NLI datasets performs the best, demonstrating strong NLI capabilities on this new domain. 



\subsection{NLI vs.\ similarity-based retrieval}\label{sec:res:retrieval}
We use the span-level annotations of ContractNLI to evaluate span retrieval over the premise for NLI. We find \ourmodel's NLI scores to provide a very strong retrieval signal, ranking one of the annotated spans at the top 61\% of the time. A random baseline, in comparison, achieves less than 1\%.



To compare with similarity-based retrievers, we adopt the TF-IDF baseline from \citet{koreeda-manning-2021-contractnli-dataset}, and also extract unsupervised sentence embeddings with BERT-large~\citep{devlin-etal-2019-bert} and SentenceT5-large~\citep{ni2021sentencet5}. 

Table~\ref{tab:contnli_res} shows the precision of each retriever at recall $0.8$ (P@R.8). Unsupervised NLI-based retrieval outperforms similarity-based retrievers. As discussed in \S\ref{sec:retrieval}, we conjecture that this is because some sentences in the document could be very similar to the hypothesis, but neutral towards it.
Table~\ref{tab:contract_ret_examples} shows an example of such case.
Within the supervised methods, SpanNLI performs better, perhaps due to only including single random spans for neutral cases in the sentence-pair training data, not utilizing the full document.

\subsection{Retrieve-and-classify}\label{sec:res_ret_cls}

\parhead{DocNLI.}
We test if our sentence-pair method can improve the low performance over long premises in DocNLI. We focus on examples with a short hypothesis (up to 20 tokens) and long premise (more than 512 tokens). Naively applying \ourmodel on these cases without sentence splitting leads to a low $0.21$ $F_1$ score. Using our retrieve-and-predict approach, the score increases up to $0.41$. Reranking doesn't seem to improve in this case, possibly due to the dataset's two-way classification format.

\parhead{ContractNLI.}
Table~\ref{tab:contnli_res} shows the main results on ContractNLI comparing sentence-pair zero-shot models with both kinds of supervised models. Surprisingly, the zero-shot sentence-pair models are competitive with the supervised SpanNLI model, even outperforming them in $F_1$(C). The supervised \ourmodel, trained also with ContractNLI, performs best overall. Even though its retrieval performance is still behind SpanNLI (\S\ref{sec:res:retrieval}), its final verdict on the hypothesis is better.

Reranking significantly improves the performance of zero-shot \ourmodel, increasing the average score by up to 27\%. We observe better performance with larger context ($K=5$).  For the supervised \ourmodel model, reranking provides only marginal gains, and increasing the context is not beneficial.

Overall, we see that sentence-pair models obtain very strong classification performance, even in a zero-shot setting (Table~\ref{tab:contnli_oracle}), while also providing descent retrieval capabilities. We hypothesize that the gap in the retrieval performance could be due to randomly sampling neutral spans instead of utilizing the full document. Augmenting the training set with more and better neutral examples could further close this gap.

\begin{table}[t]
\small
    \centering
        \resizebox{\linewidth}{!}{
    \begin{tabular}{l|c|c|ccc}
    \toprule
    Model & S.P. & P@R.8 & $F_1$(C) & $F_1$(E) & AVG \\
    \midrule
    Majority vote$^{\ddag}$ &  &  - & .083 & .428 & .256 \\
    \midrule
    \multicolumn{3}{l}{\underline{\textit{supervised:}}} \\
        SpanNLI-base$^{\ddag}$ & & .663 & .287 & .765 & .526\\
        SpanNLI-large$^{\ddag}$ & & \textbf{.793} & .357 & \textbf{.834} & .595 \\
        T5-large & v & .575 & .512 & .691 & .601\\ 
        \ourmodel$_{\text{+Cont.NLI}}$ & v & .580 & .521 & .754 & .637\\
        \ + Rerank$_{(K=1)}$ & v  & " & \textbf{.537}  & .741 & \textbf{.639}\\
        \ + Rerank$_{(K=5)}$ & v & " & .520 & .719 & .619 \\
     \midrule
    \multicolumn{3}{l}{\underline{\textit{zero-shot:}}} \\
        TF-IDF$^{\ddag}$ & v & .057 & - & - & -\\
        BERT emb. & v & .083& - & - & - \\
        SentenceT5 & v & .311& - & - & - \\
        \ourmodel$_{\text{-M.T}}$ & v & .397 & .261 & .551 & .406 \\
         \ourmodel$_{\text{-F.V}}$ & v & .397 & .247 & .594 & .420 \\
         \ourmodel & v & \textbf{.412} & .257 & .573 & .415 \\ 
         \ + Rerank$_{(K=1)}$ & v & " & .363 & \textbf{.659} & .511 \\
         \ + Rerank$_{(K=5)}$ & v & " & \textbf{.404} & .652 & \textbf{.528} \\
     \bottomrule
    \end{tabular}
    }%
    \vspace{-5pt}
    \caption{Evidence retrieval and classification results of both supervised and zero-shot models on ContractNLI test set. Within sentence-pair (S.P.) models, NLI-based retrieval is more precise than similarity retrievers. Supervised S.P.\ models  outperform the joint SpanNLI model in the final classification task thanks to better $F_1$(C). $^{\ddag}$Results from \citet{koreeda-manning-2021-contractnli-dataset}.}
    \label{tab:contnli_res}
    \vspace{-8pt}
\end{table}

\parhead{\summac.}
For evaluating on \summac, we first adopt the zero-shot method of \citet{SummaC} ($\summac_{\mathrm{ZS}}$) and only replace the backbone NLI model with \ourmodel. This improves the performance by absolute $1.2$ points (see Table~\ref{tab:summac}). This overall improvement comes despite a drop of almost 10 points on the Polytope dataset. This could be due to \summac's labeling function that treats summaries with any added or omitted content, compared to the reference, as ``factual inconsistent''. These errors relate more to the summary quality rather than its correctness and therefore, we do not expect zero-shot NLI models to catch them.

Reranking (K=1) is effective for most datasets. The best overall performance is achieved by reranking and hard aggregating the hypothesis sentences, improving over soft aggregation in 4 out of the 6 datasets, and allowing an overall $0.2$ points gain.

\subsection{Factual discrepancies in clusters}\label{sec:res_wiki}
Table~\ref{tab:wiki_corrupt} reports the results on our \textbf{Corrupted Wiki Clusters} dataset (\S\ref{sec:wiki_clusters}). 
In addition to our main method, we tried to reverse rank the spans by entailment score, but find it to perform even worst than random. We find this to be caused by pairings that support unmodified facts in the corrupted sentence. When ranking by contradiction, \ourmodel performs the best and successfully flags 68\% of the corruptions as its top prediction.

Exploring popular \textbf{Real} Wikipedia articles, without any simulated edits our known discrepancies, our method quickly identified existing inconsistencies. We attach examples in Appendix~\ref{sec:wiki_examples} and discuss them briefly here.
As depicted in Figure~\ref{fig:intro}, we find the French Wikipedia to disagree with other versions on the nationality of Bram Stoker. \ourmodel ranked this sentence highest among the 53  sentences of that cluster.
Investigating the page's history, this claim was introduced by an edit\footnote{\href{https://fr.wikipedia.org/w/index.php?title=Dracula&diff=133445656&oldid=133334932}{fr.wikipedia.org/w/index.php?title=Dracula}} in Jan.\ 2017 and remained unchanged for over 5 years.

Interestingly, when looking for consensus and ranking sentences by agreement, \ourmodel returns the English version that avoids stating any nationality: \emph{``Dracula is a novel by Bram Stoker, published in 1897.''} Intuitively, shorter statements have higher chance of obtaining consensus.

In another example, we look at the articles about ``Big Ben''. The top discrepancy prediction was a sentence from the Chinese version that was likely mistranslated due to multiple segmentation options. While this does not necessarily reveal a mistake in the original document, it shows the potential of this approach for flagging translation mistakes when related sources are available. 
The second ranked sentence identifies a statement from the Swedish page regarding the monument's official name. Upon manual verification, even though articles discuss several names that were changed over time, none seem to directly support that claim.

\begin{table}[t]
\small
    \centering
    \begin{tabular}{l|ccc}
    \toprule
    Model & \multicolumn{3}{c}{Accuracy@K} \\
    & K=1 & 5 & 10 \\
    \midrule
    Random & 17.11 & 46.53 & 73.68 \\
    \midrule
     \multicolumn{4}{l}{\underline{\textit{Reversed ranking by entailment:}}} \\
     \ourmodel &1.46 & 17.35 & 40.29 \\
    \midrule
     \multicolumn{4}{l}{\underline{\textit{Ranking by contradiction:}}} \\
    \ourmodel$_{\text{-M.T}}$ & 35.32 & 67.60 & 83.37 \\
     \ourmodel$_{\text{-F.V}}$ & 38.59 & 70.39 & 83.13 \\
     \ourmodel & \textbf{68.20} & \textbf{89.08} & \textbf{95.15} \\
     \bottomrule
    \end{tabular}
    \caption{Wikipedia corruption detection results by different ranking methods/ models. Accuracy of including the corrupted sentence within the top K predictions.}
    \label{tab:wiki_corrupt}
\end{table}

%% file: sections/related.tex
\section{Related Work}\label{sec:related}
As mentioned in the introduction, the NLI task~\citep{dagan2006, dagan2013recognizing}, sometimes called Recognizing Textual Entailment (RTE), was extensively studied by the NLP community over the past several years as a semantic reasoning benchmark~\cite[see][for surveys]{poliak-2020-survey, storks2019recent}. The field of fact verification~\citep{vlachos-riedel-2014-fact} also recently gained increased attention~\citep{Bekoulis, kotonya-toni-2020-explainable, Guo2022ASO, zeng2021automated}, sharing similar pair-wise semantic inference challenges, together with evidence retrieval.
While both tasks were found to be vulnerable to idiosyncrasies~\citep{gururangan-etal-2018-annotation,mccoy-etal-2019-right,poliak-etal-2018-hypothesis, schuster-etal-2019-towards}, methods and datasets for reducing the bias were proposed~\citep{belinkov2019, karimi-mahabadi-etal-2020-end, Shah_Schuster_Barzilay_2020, utama-etal-2020-towards, utama-etal-2021-avoiding, Yuxiang2022}.



Recently, NLI-style models were expanded for concrete purposes beyond benchmarking. For example, showing promising potential in verifying the factual correctness of dialog~\citep{gupta2021dialfact, honovich-etal-2021-q2}, summarization~\citep{chen-etal-2021-improving, eyal-etal-2019-question, alex2021qafacteval, SummaC}, and QA~\citep{bulian2022tomayto, chen-etal-2021-nli-models, mishra-etal-2021-looking} systems.
The NLI format was also found helpful for general self-training~\citep{vu-etal-2021-strata}. 
Here, we focus on real-world direct applications such as automatic contract analysis~\citep{koreeda-manning-2021-contractnli-dataset} and identifying discrepancies in document collections.

In parallel work, \citet{utama-etal-2022-falsesum} improve NLI models for evaluating summaries by generating in-domain data with automatic perturbations to simulate contradictions. We observe similar improvements when training the backbone NLI models on in-domain data for ContractNLI (supervised vs.\ zero-shot setting). Any additional improvements to the NLI model, aimed towards the target domain, are likely to further improve the reasoning capabilities of the retrieve-and-classify pipeline.



%% file: sections/conclusion.tex
\section{Conclusion}\label{sec:conclusion}
We present a comprehensive study on the performance of sentence-pair NLI models in real world applications that often involve both a shift in domain and long texts. Our findings indicate the readiness of these models to provide meaningful signal on the semantic relation between texts that can be easily aggregated towards practical gains.
To demonstrate this, we also defined a new zero-shot entailment-focused task over clusters of related documents. Our multi-task sentence-pair NLI model (\ourmodel) successfully flags spans that stand out due to their claims.

Ultimately, this study should help practitioners interested in applying NLI-style inference in real-world applications to design the best model for their target task. Our results suggest that if the hypothesis is short and the premise fully fits in the input limit of the model, a regular cross-attention classifier is likely to perform well in terms of accuracy as it is able to contextualize sentences in the premise. However, if we want to interpret the prediction by identifying the exact piece from the premise that led to the predicted conclusion, breaking the premise into segments could be useful. Furthermore, if the premise is too long to fit in the model's receptive field, then breaking the premise into segments and aggregating with our proposed techniques (retrieve-and-classify and reranking) is beneficial. Finally, when we don't have a well defined hypothesis, one might still want to automatically reason over a pair or collection of documents and identify any statements that stand out. In this case,
our reasoning over clusters methodology shows how to use strong sentence-pair classifiers to obtain useful signals that are then aggregated to highlight specific claims.

It's important to note that when designing the methods for this work, we preferred simplicity over performance in order to directly study the quality of the \ourmodel's scores. Yet, we achieve high zero-shot performance and even reach state-of-the-art on ContractNLI. We hope that this work will motivate future research on further expanding these methods, for example by decontextualizing the premise, supporting multi-hop reasoning, expanding the context with sliding windows instead of sentence splitting, or hypothesis fragmentation. 





%% file: sections/app_technical.tex
\section{Implementation details for \ourmodel} \label{sec:technical_det}

We use the T5X framework~\citep{roberts2022scaling} to finetune the T5.1.1-Large model for 500K steps and pick the best performning checkpoint on the evaluation splits of the source datasets. We use a batch size of 128 with a balanced sampling across the training datasets to account for their different sizes.

%% file: sections/app_binary.tex
\section{Zero-shot binary classification}\label{sec:binary_aggr}

DocNLI uses only two classes, ``entail'' vs.\ ``not entail'', by merging the ``neutral'' and ``contradiction'' definitions. 
The ``not entail'' instances are created with both rule-based and LM-based local perturbations over the positive pairs. This results in a slightly different task definition than NLI since even altered (i.e., ``fake'') texts can still factually agree with the hypothesis~\citep{schuster-etal-2020-limitations}. \citet{yin-etal-2021-docnli} account for this by augmenting the training set, but zero-shot NLI models might be affected by this provenance-based rather than factual-based partition of the test set.

Since NLI models are commonly trained with three target labels, adjusting to a two-way classification requires some modifications. Being a zero-shot setting, we cannot train the model's internal representations to adjust to this new label space. Instead, we can define an aggregation method. We experiment with the following variants:
\begin{enumerate}[leftmargin=*]
    \item \textbf{Entailment threshold}:  predicting ``entail'' if $p_e > T$, else ``not entail''.
    \item \textbf{Contradiction threshold}:  predicting ``not entail'' if $p_c > T$, else ``entail''.
    \item \textbf{Binary softmax}:  recompute the softmax without $s_n$, and predict ``entail'' if $\softmax(s_e, s_c)[e] > T$, else ``not entail''. 
\end{enumerate}
$T$ is a decision threshold that we can either set to some arbitrary value such as $0.5$, or calibrate it on a small set of labeled data.

Table~\ref{tab:docnli_2_aggr} presents the performance of the three aggregation methods with or without tuning $T$ on 500 random examples for the development set (with $0.05$ intervals). Thresholding on $p_e$ performs best and gives higher precision compared to the binary softmax that discards the 'neutral' score. \
Yet, all three options perform well, with different trade-offs between precision and recall, motivating the use of simple heuristics in the absence of supervised data for the target task. In the following experiments, we use the 'e' threshold method.
Tuning $T$ significantly improves the $F_1$ score by sacrificing recall for precision. We find the optimal threshold to be $0.95$, meaning that we predict 'entail' only when the model is highly confident in this relation.

\begin{table*}[t]
    \centering
    \small
    \resizebox{0.85\linewidth}{!}{
    \begin{tabular}{lc|ccc|ccc|ccc}
    \toprule
    \multicolumn{2}{c}{}& \multicolumn{3}{c}{\underline{0.2\% of DocNLI Dev.}} & \multicolumn{3}{c}{\underline{DocNLI Dev.}} & \multicolumn{3}{c}{\underline{DocNLI Test}} \\
    Aggregation & \multicolumn{1}{c}{$T$} & \multicolumn{1}{c}{Prec.} &  \multicolumn{1}{c}{Recall} &  \multicolumn{1}{c}{$F_1$(E)} & \multicolumn{1}{c}{Prec.} &  \multicolumn{1}{c}{Recall} &  \multicolumn{1}{c}{$F_1$(E)} & \multicolumn{1}{c}{Prec.} &  \multicolumn{1}{c}{Recall} &  \multicolumn{1}{c}{$F_1$(E)}\\
    \midrule
        'e' threshold & 0.5 & .198 & .746 & .313 & .216 & .810 & .341 & .220 & .800 & .345\\
                    & tuned & .319 & .610 & .\textbf{419} & .312 & .594 & .\textbf{409} & .313 & .589 & .\textbf{408}\\
        \midrule
        'c' threshold & 0.5 & .187 & .966 & .313 & .182 & .969 & .306 & .182 & .969 & .306 \\
                    & tuned & .244 & .915 & .386 & .230 & .872 & .364 & .230 & .872 & .364\\
        \midrule
        bin.\ softmax & 0.5 & .195 & .966 & .324 & .192 & .949 & .319& .193 & .949 & .321\\
            & tuned & .273 & .847 & .413 & .263 & .806 & .397 & .263 & .807 & .397 \\
                    
        \bottomrule
    \end{tabular}
    }%
    \caption{Different aggregation methods for converting predictions of a three-way NLI model to a binary label space. Aggregating by the score of the entailment class performs best. Tuning the threshold on random 500 examples (0.2\% of full Dev.) further improves the precision and $F_1$ scores, compared to the naive $0.5$ baseline.}
    \label{tab:docnli_2_aggr}
\end{table*}

%% file: sections/app_retrieval.tex
\section{Example of NLI vs.\ similarity-based retrieval}
Table~\ref{tab:contract_ret_examples} shows a retrieval example from the ContractNLI dataset, using either SentenceT5 (similarity-based) or \ourmodel (NLI-based) to retrieve spans that might support or refute the candidate hypothesis.

\begin{table*}[t]
\centering
\begin{small}
        \resizebox{\linewidth}{!}{
\begin{tabular}{p{0.09\linewidth}|p{0.9\linewidth}}
\toprule
Hypothesis: & Confidential Information shall only include technical information. $\qquad$ $\qquad$ $\qquad$ \emph{(gold label = contradiction)} \\
\midrule
ST5 top-1 & 5.1.2. use Confidential Information only for the Project;\\
\cmidrule{2-2}
top-2 & 5.3.2. The disclosure of Confidential Information to Recipient or its Representatives shall not give Recipient or its Representatives any licence or other rights in relation to that Confidential Information [...]\\
\midrule
\ourmodel argmax $p_c$ & 3.5. "Confidential information" means any information of whatever form relating to the Project or Discloser or any of its Affiliates or Clients, supplied or made available by Discloser or on its behalf to recipient [...] \\
\cmidrule{2-2}
argmax $p_e$ & You the subject-matter expert \\
\bottomrule
\end{tabular}
}%
\caption{ContractNLI evidence retrieval example. The top two retrievals of the SentenceT5 (ST5) model relate to the hypothesis (discussing confidential information), but are do not refute or support it. Alternatively, retrieving by NLI scores is highly informative as the  sentence with max $p_c$ clearly contradicts the hypothesis.}\label{tab:contract_ret_examples}
\end{small}
\end{table*}

%% file: sections/app_dataset.tex
\section{\summac Evaluation}\label{app:summac}

\begin{table*}[t]
\small
    \centering
    \begin{tabular}{l|dddddd|d}
    \toprule
    Method & \multicolumn{1}{c}{CGS} & \multicolumn{1}{c}{XSF} & \multicolumn{1}{c}{poly}& \multicolumn{1}{c}{factC} & \multicolumn{1}{c}{sumEv}& \multicolumn{1}{c}{frank} & \multicolumn{1}{c}{AVG}\\
    \midrule
    $\summac_{\mathrm{ZS}}$ & 70.4 & 58.4 & \textbf{62}.\textbf{0} & 83.8 & 78.7 & 79.0 & 72.1 \\
    \midrule
    \ourmodel  (soft) & 79.3 & 59.3 & 52.4 & \textbf{89}.\textbf{5} & 77.2 & \textbf{82}.\textbf{1} & 73.3 \\
    + Rerank (soft) & 79.6 & 62.7 & 52.8 & 86.1 & 78.5 & 80.4 & 73.3 \\
    + Rerank (hard)  & \textbf{80}.\textbf{5} & \textbf{64}.\textbf{2} & 55.1 & 83.3 & \textbf{79}.\textbf{7} & 78.4 & \textbf{73}.\textbf{5} \\ 
     \bottomrule
    \end{tabular}
    \caption{\summac zero-shot balanced accuracy. The hypothesis aggregation method (\S\ref{sec:multi_sent_hyp}) is in parenthesis.}
    \label{tab:summac}
\end{table*}

The \summac benchmark includes six datasets: CGS~\citep{falke-etal-2019-ranking},
XSF~\citep{maynez-etal-2020-faithfulness},
Polytope~\citep{huang-etal-2020-achieved},
FactCC~\citep{kryscinski-etal-2020-evaluating},
SummEval~\citep{fabbri-etal-2021-summeval},
Frank~\citep{pagnoni-etal-2021-understanding}.

Table~\ref{tab:summac} reports the results on this benchmark. See \S\ref{sec:res_ret_cls} for discussion.

\section{Additional Details on the Wiki Clusters Dataset}\label{sec:wikiclusters_details}

As described in Section~\ref{sec:wiki_clusters}, we created a Wikipedia-based dataset with clusters of similar articles written by different communities in multiple languages. Below, we provide additional details on the process.

We first collect the 5000 most popular accessed pages in the English Wikipedia, according to the ranking of November 2021.\footnote{\url{https://bit.ly/Wiki_popular_pages}}
We take a cleaned version (without links) of the introduction of each page (the text coming before the content table). Then, to create the cluster for each article, we use each article's language links and collect similar introductions from the 10 (non-EN) languages that with most admins as of November 2021.\footnote{By \url{https://en.wikipedia.org/wiki/List_of_Wikipedias\#Edition_details}: DE, FR, IT, PL, RU, SV, ZH, ES, PT, UK.}
Finally, we use Google Translate API to translate all articles to English. 

We manually examine a random subset of the clusters and find very few translation mistakes and that in most clusters there are at least several non-English introductions with sufficient length and content. This is mostly thanks to our choice of popular pages and languages. We also tried to randomly sample English articles but find many of the versions in other languages to be missing or have a single sentence.

As mentioned in Section~\ref{sec:wiki_clusters}, we designed a controlled experiment where we simulate local corruptions to the English version, and use the information from the other versions to predict which sentence was changed. To make the corruptions realistic and not obvious, we rely on the Wikipedia edits collection from the VitaminC test set~\cite{schuster-etal-2021-get}. These edits include both real revisions from Wikipedia history logs, and synthetic edits created by annotators to modify certain facts. To ensure that we only use edits that present a factual modification, we only take pairs of evidence from the same page that are paired with the same claim, but express an opposing relation (one supports it and the other refutes it).

To match the edits from VitaminC with our current version, we use word-level (also splitting hyphens) Jaccard similarity. First, looking from small edits, we only take the VitaminC edit instances with similarity greater than $0.25$ between the ``before'' ($x_{b}$) and ``after'' ($x_{a}$) sentences. Then, for each of the edits, we look for the sentence $x$ from the current article that has the highest Jaccard similarity with either the $x_{b}$ or $x_{a}$ sentence. If no sentence has greater than $0.2$ similarity, we skip this edit. Finally, to decide which of the two sentences is coherent with the current version, we pick the one with the higher Jaccard similarity with $x$. Accordingly, we assume that the other sentence represent a factual modification to the current article, and therefore create a local discrepancy by replacing it with $x$.

Following this process, we obtain a total of 824 local corruptions to 144 different articles. See Table~\ref{tab:wiki_cur_example} for an example.

\begin{table*}[ht]
\centering
\begin{small}
        \resizebox{\linewidth}{!}{
\begin{tabular}{p{0.1\linewidth}|p{0.9\linewidth}}
\toprule
Original & Mars is the site of Olympus Mons, the largest volcano and highest known mountain on any planet in the Solar System, and of Valles Marineris, one of the largest canyons in the Solar System. \\
Corruption & Mars is the site of Olympus Mons, the largest volcano and second-highest known mountain in the Solar System, but far away from Valles Marineris, one of the largest canyons in the Solar System.\\

\midrule
\multicolumn{2}{l}{\emph{Examples of related sentences from other articles with helpful information for identifying the corruption:}} \\
FR & The highest volcano in the Solar System, Olympus Mons (which is a shield volcano), and the largest canyon, Valles Marineris, are found on Mars. \\
IT & Among the most noteworthy geological formations of Mars are: Olympus Mons, or Mount Olympus, the largest volcano in the solar system (27 km high); the Valles Marineris, a long canyon considerably larger than the terrestrial ones; and a huge crater on the northern hemisphere, about 40\% wide of the entire Martian surface. \\
SV & During large parts of Mars' history, long-lasting volcanic eruptions occurred which, among other things, created Olympus Mons, the highest mountain in the solar system.\\

\bottomrule
\end{tabular}
}%
\caption{Example of a \emph{simulated} corruption in the English Wikipedia about Mars from our corrupted articles dataset (\S\ref{sec:wiki_clusters}). We are given the introduction of the English article, consisting of 35 sentences, where one sentence, the ``original'', was replaced with the ``corruption'' one. The goal is to successfully identify which sentence was corrupted by leveraging information from the other 10 related articles (each with 18 sentences on average). We only present the most relevant sentences here, but the model has to read through the whole cluster.}\label{tab:wiki_cur_example}
\end{small}
\end{table*}

%% file: sections/app_wiki.tex
\section{Examples of discrepancies and consensus in Wikipedia}\label{sec:wiki_examples}

We present the retrievals of our method for identifying discrepancies and consensus in document clusters (\S\ref{sec:wiki_clusters}) when applied on the ``Dracula'' (Tables~\ref{tab:dracula_dis}-\ref{tab:dracula_con}), ``Big Ben'' (Tables~\ref{tab:bigben_dis}-\ref{tab:bigben_con}), and ``Cameron Boyce'' (Tables~\ref{tab:cameron_dis}-\ref{tab:cameron_con}) pages.
We observe that in general shorter sentences with consensus tend to be short and concise. This is intuitive as longer sentence are more likely to include claims that are missing from other documents.

\begin{table*}[ht]
\centering
\begin{small}
        \resizebox{\linewidth}{!}{
\begin{tabular}{p{0.05\linewidth}|p{0.95\linewidth}}
\multicolumn{2}{l}{\emph{Dracula: searching for \textbf{discrepancies}.}} \\
\toprule
FR & Dracula is an epistolary novel by British writer Bram Stoker published in 1897. \\
\midrule
\multicolumn{2}{l}{\emph{Top sentence from each document by disagreement with the candidate:}} \\
EN & A small group, led by Abraham Van Helsing, hunt Dracula and, in the end, kill him.\\
DE & Dracula is a novel by Irish writer Bram Stoker published in 1897.\\
IT & Dracula is an epistolary novel written by Irish Bram Stoker in 1897, inspired by the figure of Vlad III, prince of Wallachia, and is one of the last examples of Gothic novels.\\
PL & Dracula - a 19th-century Gothic novel by the Irish writer Bram Stoker, depicting the fight of a group of volunteers with the vampire Dracula.\\
RU & Dracula is a novel by the Irish writer Bram Stoker, first published in 1897.\\
PT & Dracula (Dracula) is an 1897 gothic horror novel written by Irish author Bram Stoker, starring the vampire Count Dracula.\\
ES & Dracula is a novel published in 1897 by the Irishman Bram Stoker, as a result of which his antagonist character, Count Dracula, became the quintessential Western vampire archetype, becoming considered the most famous vampire.\\
ZH & "Dracula" is a gothic horror novel based on vampires published in 1897 by Irish writer Bram Stoker.\\
SV & Dracula is a horror novel from 1897 by the Irish author Bram Stoker, in which the main antagonist is the vampire Count Dracula.\\
UK & Dracula is a novel by Irish writer Bram Stoker, first published in 1897.\\
\bottomrule
\end{tabular}
}%
\caption{The sentence with highest discrepancy score (shown at the top) among all 53 sentences from ``Dracula'' articles. Beneath, we show the sentence from each Wikipedia version that had the highest disagreement score with the candidate. This example is also illustrated in Figure~\ref{fig:intro}.}\label{tab:dracula_dis}
\end{small}
\end{table*}

\begin{table*}[ht]
\centering
\begin{small}
        \resizebox{\linewidth}{!}{
\begin{tabular}{p{0.05\linewidth}|p{0.95\linewidth}}
\multicolumn{2}{l}{\emph{Dracula: searching for \textbf{consensus}.}} \\
\toprule
EN & Dracula is a novel by Bram Stoker, published in 1897. \\
\midrule
\multicolumn{2}{l}{\emph{Top sentence from each document by agreement with the candidate:}} \\
DE & Dracula is a novel by Irish writer Bram Stoker published in 1897. \\
FR & Dracula is an epistolary novel by British writer Bram Stoker published in 1897. \\
IT & Dracula is an epistolary novel written by Irish Bram Stoker in 1897, inspired by the figure of Vlad III, prince of Wallachia, and is one of the last examples of Gothic novels.\\
PL & Dracula - a 19th-century Gothic novel by the Irish writer Bram Stoker, depicting the fight of a group of volunteers with the vampire Dracula.\\
RU & Dracula is a novel by the Irish writer Bram Stoker, first published in 1897.\\
PT & Dracula (Dracula) is an 1897 gothic horror novel written by Irish author Bram Stoker, starring the vampire Count Dracula.\\
ES & Dracula is a novel published in 1897 by the Irishman Bram Stoker, as a result of which his antagonist character, Count Dracula, became the quintessential Western vampire archetype, becoming considered the most famous vampire.\\
ZH & "Dracula" is a gothic horror novel based on vampires published in 1897 by Irish writer Bram Stoker.\\
SV & Dracula is a horror novel from 1897 by the Irish author Bram Stoker, in which the main antagonist is the vampire Count Dracula.\\
UK & Dracula is a novel by Irish writer Bram Stoker, first published in 1897.\\
\bottomrule
\end{tabular}
}%
\caption{The sentence with the most consensus (shown at the top) among all 53 sentences from ``Dracula'' articles. Beneath, we show the sentence from each Wikipedia version that had the highest agreement score with the candidate.}\label{tab:dracula_con}
\end{small}
\end{table*}

\begin{table*}[ht]
\centering
\begin{small}
        \resizebox{\linewidth}{!}{
\begin{tabular}{p{0.05\linewidth}|p{0.95\linewidth}}
\multicolumn{2}{l}{\emph{Big Ben: searching for \textbf{discrepancies}.}} \\
\toprule
SV & Big Ben is officially called the Great Bell of Westminster and strikes every hour in the tower clock with the official name Great Clock of Westminster. \\
\midrule
\multicolumn{2}{l}{\emph{Top sentence from each document by disagreement with the candidate:}} \\
EN & The official name of the tower in which Big Ben is located was originally the Clock Tower, but it was renamed Elizabeth Tower in 2012, to mark the Diamond Jubilee of Elizabeth II, Queen of the United Kingdom.\\
DE & The tower has been officially called Elizabeth Tower since September 2012.\\
FR & Previously, it was simply called the Clock Tower.\\
IT & This bell tower rings every quarter of an hour.\\
PL & On September 12, 2012, the tower was officially named Elizabeth Tower in honor of Elizabeth II's 60-year reign.\\
RU & The official name of the tower since 2012 is the Elizabeth Tower, one of the most recognizable symbols of Great Britain, often used in souvenirs, advertisements, and movies.\\
PT & The official name of the tower in which Big Ben is located was originally Clock Tower, but it was renamed Elizabeth Tower in 2012 to mark Queen Elizabeth II's Diamond Jubilee.\\
ES & Its official name was Clock Tower, until on June 26, 2012, in honor of Queen Elizabeth II's Diamond Jubilee, it was decided that the tower would be renamed Elizabeth Tower.\\
ZH & Big Ben (English: Big Ben, or translated as Big Ben) is a big newspaper clock located at the north end of the Palace of Westminster in London.\\
UK & The official name of the tower since 2012 - Elizabeth Tower (English Elizabeth Tower).\\
\bottomrule
\end{tabular}
}%
\caption{The sentence with second highest discrepancy score (shown at the top) among all 72 sentences from ``Big Ben'' articles. Beneath, we show the sentence from each Wikipedia version that had the highest disagreement score with the candidate.}\label{tab:bigben_dis}
\end{small}
\end{table*}

\begin{table*}[ht]
\centering
\begin{small}
        \resizebox{\linewidth}{!}{
\begin{tabular}{p{0.05\linewidth}|p{0.95\linewidth}}
\multicolumn{2}{l}{\emph{Big Ben: searching for \textbf{consensus}.}} \\
\toprule
UK & The official name of the tower since 2012 - Elizabeth Tower (English Elizabeth Tower). \\
\midrule
\multicolumn{2}{l}{\emph{Top sentence from each document by agreement with the candidate:}} \\
EN & The official name of the tower in which Big Ben is located was originally the Clock Tower, but it was renamed Elizabeth Tower in 2012, to mark the Diamond Jubilee of Elizabeth II, Queen of the United Kingdom.\\
DE & The tower has been officially called Elizabeth Tower since September 2012.\\
FR & The tower was renamed on the occasion of the Diamond Jubilee of Elizabeth II in 2012.\\
IT & Known as the Clock Tower, the name was officially changed to Elizabeth Tower on the occasion of Elizabeth II's Diamond Jubilee in June 2012.\\
PL & On September 12, 2012, the tower was officially named Elizabeth Tower in honor of Elizabeth II's 60-year reign.\\
RU & The official name of the tower since 2012 is the Elizabeth Tower, one of the most recognizable symbols of Great Britain, often used in souvenirs, advertisements, and movies.\\
PT & The official name of the tower in which Big Ben is located was originally Clock Tower, but it was renamed Elizabeth Tower in 2012 to mark Queen Elizabeth II's Diamond Jubilee.\\
ES & Its official name was Clock Tower, until on June 26, 2012, in honor of Queen Elizabeth II's Diamond Jubilee, it was decided that the tower would be renamed Elizabeth Tower.\\
ZH & Big Ben (English: Big Ben, or translated as Big Ben) is a big newspaper clock located at the north end of the Palace of Westminster in London.\\
SV & The tower, which has been called Elizabeth Tower since 2012 and where the clock hangs, is 96.3 meters high.\\
\bottomrule
\end{tabular}
}%
\caption{The sentence with the most consensus (shown at the top) among all 72 sentences from ``Big Ben'' articles. Beneath, we show the sentence from each Wikipedia version that had the highest agreement score with the candidate.}\label{tab:bigben_con}
\end{small}
\end{table*}

\begin{table*}[ht]
\centering
\begin{small}
        \resizebox{\linewidth}{!}{
\begin{tabular}{p{0.03\linewidth}|p{0.98\linewidth}}
\multicolumn{2}{l}{\emph{Cameron Boyce: searching for \textbf{discrepancies}.}} \\
\toprule
UK & He died of an epileptic seizure on the night of July 7, 2019 at 2:35 p.m. \\
\midrule
\multicolumn{2}{l}{\emph{Top sentence from each document by disagreement with the candidate:}} \\
ZH & On July 6, 2019, Boyce died of epileptic seizures at the age of 20.\\
RU & Cameron Mica Boyce (born May 28, 1999 - July 6, 2019) - American actor and dancer, best known for his leading roles in the comedy series Jesse (2011-2015) and Gamer's Diary (2015-2017) , as well as in the series of films "Descendants" (2015-2019).\\
EN & Cameron Mica Boyce (May 28, 1999 – July 6, 2019) was an American actor.\\
PL & Cameron Mica "Cam" Boyce (born May 28, 1999 in Los Angeles, died July 6, 2019 therein) - American actor, dancer and model.\\
FR & Cameron Boyce is an American actor, dancer, singer and model, born May 28, 1999 in Los Angeles (California) and died July 6, 2019 in the same city.\\
ES & Cameron Mica Boyce (Los Angeles, California; May 28, 1999-July 6, 2019) was an American actor known primarily for his roles in the feature films Descendants, Descendants 2, Descendants 3, as well as for his role of Luke Ross on the Disney Channel series Jessie.\\
PT & Cameron Boyce (Los Angeles, May 28, 1999 – Los Angeles, July 6, 2019) was an American actor, singer, dancer and voice actor, known for starring in films such as Mirrors and appearing in Eagle Eye, both of 2008.\\
SV & Cameron Boyce, born May 28, 1999 in Los Angeles, died July 6, 2019 in Los Angeles, was an American actor.\\
DE & Cameron Boyce (born May 28, 1999 in Los Angeles, California - July 6, 2019) was an American actor and child actor.\\
IT & Cameron Mica Boyce (Los Angeles, May 28, 1999 - Los Angeles, July 6, 2019) was an American actor and dancer.\\
\bottomrule
\end{tabular}
}%
\caption{The sentence with highest discrepancy score (shown at the top) among all 38 sentences from ``Cameron Boyce'' articles. Beneath, we show the sentence from each Wikipedia version that had the highest disagreement score with the candidate.}\label{tab:cameron_dis}
\end{small}
\end{table*}

\begin{table*}[ht]
\centering
\begin{small}
        \resizebox{\linewidth}{!}{
\begin{tabular}{p{0.03\linewidth}|p{0.98\linewidth}}
\multicolumn{2}{l}{\emph{Cameron Boyce: searching for \textbf{consensus}.}} \\
\toprule
EN & Cameron Mica Boyce (May 28, 1999 – July 6, 2019) was an American actor. \\
\midrule
\multicolumn{2}{l}{\emph{Top sentence from each document by agreement with the candidate:}} \\
PL & Cameron Mica "Cam" Boyce (born May 28, 1999 in Los Angeles, died July 6, 2019 therein) - American actor, dancer and model.\\
RU & Cameron Mica Boyce (born May 28, 1999 - July 6, 2019) - American actor and dancer, best known for his leading roles in the comedy series Jesse (2011-2015) and Gamer's Diary (2015-2017) , as well as in the series of films "Descendants" (2015-2019).\\
ES & Cameron Mica Boyce (Los Angeles, California; May 28, 1999-July 6, 2019) was an American actor known primarily for his roles in the feature films Descendants, Descendants 2, Descendants 3, as well as for his role of Luke Ross on the Disney Channel series Jessie.\\
IT & Cameron Mica Boyce (Los Angeles, May 28, 1999 - Los Angeles, July 6, 2019) was an American actor and dancer.\\
ZH & Cameron Mica Boyce (English: Cameron Mica Boyce, May 28, 1999-July 6, 2019) was an American actor and dancer.\\
PT & Cameron is best known for playing Luke Ross in Jessie, a series that aired on the Disney Channel, and for his role in the film Descendants, having also played the role of Conor in Gamer's Guide to Pretty Much Everything.\\
SV & Boyce was known for his roles in films such as Mirrors, Eagle Eye, Grown Ups and Grown Ups 2.\\
FR & Cameron Boyce is an American actor, dancer, singer and model, born May 28, 1999 in Los Angeles (California) and died July 6, 2019 in the same city.\\
DE & Cameron Boyce (born May 28, 1999 in Los Angeles, California - July 6, 2019) was an American actor and child actor.\\
UK & Cameron Boyce is an American actor and dancer best known for his roles in the feature films "Mirrors," "Hook," "Classmates," "Heirs," "Heirs 2" and the Disney series Jesse.\\
\bottomrule
\end{tabular}
}%
\caption{The sentence with the most consensus (shown at the top) among all 38 sentences from ``Cameron Boyce'' articles. Beneath, we show the sentence from each Wikipedia version that had the highest agreement score with the candidate.}\label{tab:cameron_con}
\end{small}
\end{table*}

%% file: main.bbl
\begin{thebibliography}{58}
\expandafter\ifx\csname natexlab\endcsname\relax\def\natexlab#1{#1}\fi

\bibitem[{Aribandi et~al.(2022)Aribandi, Tay, Schuster, Rao, Zheng, Mehta,
  Zhuang, Tran, Bahri, Ni, Gupta, Hui, Ruder, and Metzler}]{arib2021ext5}
Vamsi Aribandi, Yi~Tay, Tal Schuster, Jinfeng Rao, Huaixiu~Steven Zheng,
  Sanket~Vaibhav Mehta, Honglei Zhuang, Vinh~Q. Tran, Dara Bahri, Jianmo Ni,
  Jai Gupta, Kai Hui, Sebastian Ruder, and Donald Metzler. 2022.
\newblock \href {https://openreview.net/forum?id=Vzh1BFUCiIX} {Ext5: Towards
  extreme multi-task scaling for transfer learning}.
\newblock In \emph{International Conference on Learning Representations}.

\bibitem[{Bekoulis et~al.(2021)Bekoulis, Papagiannopoulou, and
  Deligiannis}]{Bekoulis}
Giannis Bekoulis, Christina Papagiannopoulou, and Nikos Deligiannis. 2021.
\newblock \href {https://doi.org/10.1145/3485127} {A review on fact extraction
  and verification}.
\newblock \emph{ACM Comput. Surv.}, 55(1).

\bibitem[{Belinkov et~al.(2019)Belinkov, Poliak, Shieber, Van~Durme, and
  Rush}]{belinkov2019}
Yonatan Belinkov, Adam Poliak, Stuart Shieber, Benjamin Van~Durme, and
  Alexander Rush. 2019.
\newblock \href {https://doi.org/10.18653/v1/s19-1028} {On adversarial removal
  of hypothesis-only bias in natural language inference}.
\newblock \emph{Proceedings of the Eighth Joint Conference on Lexical and
  Computational Semantics (*SEM 2019)}.

\bibitem[{Beltagy et~al.(2020)Beltagy, Peters, and Cohan}]{longformer}
Iz~Beltagy, Matthew~E. Peters, and Arman Cohan. 2020.
\newblock \href {http://arxiv.org/abs/2004.05150} {Longformer: The
  long-document transformer}.

\bibitem[{Bowman et~al.(2015)Bowman, Angeli, Potts, and
  Manning}]{bowman-etal-2015-large}
Samuel~R. Bowman, Gabor Angeli, Christopher Potts, and Christopher~D. Manning.
  2015.
\newblock \href {https://doi.org/10.18653/v1/D15-1075} {A large annotated
  corpus for learning natural language inference}.
\newblock In \emph{Proceedings of the 2015 Conference on Empirical Methods in
  Natural Language Processing}, pages 632--642, Lisbon, Portugal. Association
  for Computational Linguistics.

\bibitem[{Bulian et~al.(2022)Bulian, Buck, Gajewski, Boerschinger, and
  Schuster}]{bulian2022tomayto}
Jannis Bulian, Christian Buck, Wojciech Gajewski, Benjamin Boerschinger, and
  Tal Schuster. 2022.
\newblock \href {http://arxiv.org/abs/2202.07654} {Tomayto, tomahto. beyond
  token-level answer equivalence for question answering evaluation}.

\bibitem[{Chen et~al.(2021{\natexlab{a}})Chen, Choi, and
  Durrett}]{chen-etal-2021-nli-models}
Jifan Chen, Eunsol Choi, and Greg Durrett. 2021{\natexlab{a}}.
\newblock \href {https://doi.org/10.18653/v1/2021.findings-emnlp.324} {Can
  {NLI} models verify {QA} systems{'} predictions?}
\newblock In \emph{Findings of the Association for Computational Linguistics:
  EMNLP 2021}, pages 3841--3854, Punta Cana, Dominican Republic. Association
  for Computational Linguistics.

\bibitem[{Chen et~al.(2021{\natexlab{b}})Chen, Zhang, Sone, and
  Roth}]{chen-etal-2021-improving}
Sihao Chen, Fan Zhang, Kazoo Sone, and Dan Roth. 2021{\natexlab{b}}.
\newblock \href {https://doi.org/10.18653/v1/2021.naacl-main.475} {Improving
  faithfulness in abstractive summarization with contrast candidate generation
  and selection}.
\newblock In \emph{Proceedings of the 2021 Conference of the North American
  Chapter of the Association for Computational Linguistics: Human Language
  Technologies}, pages 5935--5941, Online. Association for Computational
  Linguistics.

\bibitem[{Choi et~al.(2021)Choi, Palomaki, Lamm, Kwiatkowski, Das, and
  Collins}]{choi-etal-2021-decontextualization}
Eunsol Choi, Jennimaria Palomaki, Matthew Lamm, Tom Kwiatkowski, Dipanjan Das,
  and Michael Collins. 2021.
\newblock \href {https://doi.org/10.1162/tacl_a_00377} {Decontextualization:
  Making sentences stand-alone}.
\newblock \emph{Transactions of the Association for Computational Linguistics},
  9:447--461.

\bibitem[{Dagan et~al.(2006)Dagan, Glickman, and Magnini}]{dagan2006}
Ido Dagan, Oren Glickman, and Bernardo Magnini. 2006.
\newblock The pascal recognising textual entailment challenge.
\newblock In \emph{Machine Learning Challenges. Evaluating Predictive
  Uncertainty, Visual Object Classification, and Recognising Tectual
  Entailment}, pages 177--190, Berlin, Heidelberg. Springer Berlin Heidelberg.

\bibitem[{Dagan et~al.(2013)Dagan, Roth, Sammons, and
  Zanzotto}]{dagan2013recognizing}
Ido Dagan, Dan Roth, Mark Sammons, and Fabio~Massimo Zanzotto. 2013.
\newblock Recognizing textual entailment: Models and applications.
\newblock \emph{Synthesis Lectures on Human Language Technologies},
  6(4):1--220.

\bibitem[{Devlin et~al.(2019)Devlin, Chang, Lee, and
  Toutanova}]{devlin-etal-2019-bert}
Jacob Devlin, Ming-Wei Chang, Kenton Lee, and Kristina Toutanova. 2019.
\newblock \href {https://doi.org/10.18653/v1/N19-1423} {{BERT}: Pre-training of
  deep bidirectional transformers for language understanding}.
\newblock In \emph{Proceedings of the 2019 Conference of the North {A}merican
  Chapter of the Association for Computational Linguistics: Human Language
  Technologies, Volume 1 (Long and Short Papers)}, pages 4171--4186,
  Minneapolis, Minnesota. Association for Computational Linguistics.

\bibitem[{Eyal et~al.(2019)Eyal, Baumel, and Elhadad}]{eyal-etal-2019-question}
Matan Eyal, Tal Baumel, and Michael Elhadad. 2019.
\newblock \href {https://doi.org/10.18653/v1/N19-1395} {Question answering as
  an automatic evaluation metric for news article summarization}.
\newblock In \emph{Proceedings of the 2019 Conference of the North {A}merican
  Chapter of the Association for Computational Linguistics: Human Language
  Technologies, Volume 1 (Long and Short Papers)}, pages 3938--3948,
  Minneapolis, Minnesota. Association for Computational Linguistics.

\bibitem[{Fabbri et~al.(2021{\natexlab{a}})Fabbri, Kry{\'s}ci{\'n}ski, McCann,
  Xiong, Socher, and Radev}]{fabbri-etal-2021-summeval}
Alexander~R. Fabbri, Wojciech Kry{\'s}ci{\'n}ski, Bryan McCann, Caiming Xiong,
  Richard Socher, and Dragomir Radev. 2021{\natexlab{a}}.
\newblock \href {https://doi.org/10.1162/tacl_a_00373} {{S}umm{E}val:
  Re-evaluating summarization evaluation}.
\newblock \emph{Transactions of the Association for Computational Linguistics},
  9:391--409.

\bibitem[{Fabbri et~al.(2021{\natexlab{b}})Fabbri, Wu, Liu, and
  Xiong}]{alex2021qafacteval}
Alexander~R. Fabbri, Chien-Sheng Wu, Wenhao Liu, and Caiming Xiong.
  2021{\natexlab{b}}.
\newblock \href {http://arxiv.org/abs/2112.08542} {Qafacteval: Improved
  qa-based factual consistency evaluation for summarization}.

\bibitem[{Falke et~al.(2019)Falke, Ribeiro, Utama, Dagan, and
  Gurevych}]{falke-etal-2019-ranking}
Tobias Falke, Leonardo F.~R. Ribeiro, Prasetya~Ajie Utama, Ido Dagan, and Iryna
  Gurevych. 2019.
\newblock \href {https://doi.org/10.18653/v1/P19-1213} {Ranking generated
  summaries by correctness: An interesting but challenging application for
  natural language inference}.
\newblock In \emph{Proceedings of the 57th Annual Meeting of the Association
  for Computational Linguistics}, pages 2214--2220, Florence, Italy.
  Association for Computational Linguistics.

\bibitem[{Guo et~al.(2022)Guo, Schlichtkrull, and Vlachos}]{Guo2022ASO}
Zhijiang Guo, M.~Schlichtkrull, and Andreas Vlachos. 2022.
\newblock A survey on automated fact-checking.
\newblock \emph{Transactions of the Association for Computational Linguistics},
  10:178--206.

\bibitem[{Gupta et~al.(2021)Gupta, Wu, Liu, and Xiong}]{gupta2021dialfact}
Prakhar Gupta, Chien-Sheng Wu, Wenhao Liu, and Caiming Xiong. 2021.
\newblock \href {http://arxiv.org/abs/2110.08222} {Dialfact: A benchmark for
  fact-checking in dialogue}.

\bibitem[{Gururangan et~al.(2018)Gururangan, Swayamdipta, Levy, Schwartz,
  Bowman, and Smith}]{gururangan-etal-2018-annotation}
Suchin Gururangan, Swabha Swayamdipta, Omer Levy, Roy Schwartz, Samuel Bowman,
  and Noah~A. Smith. 2018.
\newblock \href {https://doi.org/10.18653/v1/N18-2017} {Annotation artifacts in
  natural language inference data}.
\newblock In \emph{Proceedings of the 2018 Conference of the North {A}merican
  Chapter of the Association for Computational Linguistics: Human Language
  Technologies, Volume 2 (Short Papers)}, pages 107--112, New Orleans,
  Louisiana. Association for Computational Linguistics.

\bibitem[{Honovich et~al.(2021)Honovich, Choshen, Aharoni, Neeman, Szpektor,
  and Abend}]{honovich-etal-2021-q2}
Or~Honovich, Leshem Choshen, Roee Aharoni, Ella Neeman, Idan Szpektor, and Omri
  Abend. 2021.
\newblock \href {https://doi.org/10.18653/v1/2021.emnlp-main.619} {$q^{2}$:
  {E}valuating factual consistency in knowledge-grounded dialogues via question
  generation and question answering}.
\newblock In \emph{Proceedings of the 2021 Conference on Empirical Methods in
  Natural Language Processing}, pages 7856--7870, Online and Punta Cana,
  Dominican Republic. Association for Computational Linguistics.

\bibitem[{Huang et~al.(2020)Huang, Cui, Yang, Bao, Wang, Xie, and
  Zhang}]{huang-etal-2020-achieved}
Dandan Huang, Leyang Cui, Sen Yang, Guangsheng Bao, Kun Wang, Jun Xie, and Yue
  Zhang. 2020.
\newblock \href {https://doi.org/10.18653/v1/2020.emnlp-main.33} {What have we
  achieved on text summarization?}
\newblock In \emph{Proceedings of the 2020 Conference on Empirical Methods in
  Natural Language Processing (EMNLP)}, pages 446--469, Online. Association for
  Computational Linguistics.

\bibitem[{IV et~al.(2021)IV, Passos, Singh, and Chang}]{iv2021fruit}
Robert L.~Logan IV, Alexandre Passos, Sameer Singh, and Ming-Wei Chang. 2021.
\newblock \href {http://arxiv.org/abs/2112.08634} {Fruit: Faithfully reflecting
  updated information in text}.

\bibitem[{Jiang et~al.(2020)Jiang, Bordia, Zhong, Dognin, Singh, and
  Bansal}]{jiang-etal-2020-hover}
Yichen Jiang, Shikha Bordia, Zheng Zhong, Charles Dognin, Maneesh Singh, and
  Mohit Bansal. 2020.
\newblock \href {https://doi.org/10.18653/v1/2020.findings-emnlp.309}
  {{H}o{V}er: A dataset for many-hop fact extraction and claim verification}.
\newblock In \emph{Findings of the Association for Computational Linguistics:
  EMNLP 2020}, pages 3441--3460, Online. Association for Computational
  Linguistics.

\bibitem[{Karimi~Mahabadi et~al.(2020)Karimi~Mahabadi, Belinkov, and
  Henderson}]{karimi-mahabadi-etal-2020-end}
Rabeeh Karimi~Mahabadi, Yonatan Belinkov, and James Henderson. 2020.
\newblock \href {https://doi.org/10.18653/v1/2020.acl-main.769} {End-to-end
  bias mitigation by modelling biases in corpora}.
\newblock In \emph{Proceedings of the 58th Annual Meeting of the Association
  for Computational Linguistics}, pages 8706--8716, Online. Association for
  Computational Linguistics.

\bibitem[{Koreeda and Manning(2021)}]{koreeda-manning-2021-contractnli-dataset}
Yuta Koreeda and Christopher Manning. 2021.
\newblock \href {https://doi.org/10.18653/v1/2021.findings-emnlp.164}
  {{C}ontract{NLI}: A dataset for document-level natural language inference for
  contracts}.
\newblock In \emph{Findings of the Association for Computational Linguistics:
  EMNLP 2021}, pages 1907--1919, Punta Cana, Dominican Republic. Association
  for Computational Linguistics.

\bibitem[{Kotonya and Toni(2020)}]{kotonya-toni-2020-explainable}
Neema Kotonya and Francesca Toni. 2020.
\newblock \href {https://doi.org/10.18653/v1/2020.coling-main.474} {Explainable
  automated fact-checking: A survey}.
\newblock In \emph{Proceedings of the 28th International Conference on
  Computational Linguistics}, pages 5430--5443, Barcelona, Spain (Online).
  International Committee on Computational Linguistics.

\bibitem[{Kryscinski et~al.(2020)Kryscinski, McCann, Xiong, and
  Socher}]{kryscinski-etal-2020-evaluating}
Wojciech Kryscinski, Bryan McCann, Caiming Xiong, and Richard Socher. 2020.
\newblock \href {https://doi.org/10.18653/v1/2020.emnlp-main.750} {Evaluating
  the factual consistency of abstractive text summarization}.
\newblock In \emph{Proceedings of the 2020 Conference on Empirical Methods in
  Natural Language Processing (EMNLP)}, pages 9332--9346, Online. Association
  for Computational Linguistics.

\bibitem[{Laban et~al.(2022)Laban, Schnabel, Bennett, and Hearst}]{SummaC}
Philippe Laban, Tobias Schnabel, Paul~N. Bennett, and Marti~A. Hearst. 2022.
\newblock \href {https://doi.org/10.1162/tacl_a_00453} {Summac: Re-visiting
  nli-based models for inconsistency detection in summarization}.
\newblock \emph{Transactions of the Association for Computational Linguistics},
  10:163–177.

\bibitem[{Liu et~al.(2019)Liu, Ott, Goyal, Du, Joshi, Chen, Levy, Lewis,
  Zettlemoyer, and Stoyanov}]{roberta}
Yinhan Liu, Myle Ott, Naman Goyal, Jingfei Du, Mandar Joshi, Danqi Chen, Omer
  Levy, Mike Lewis, Luke Zettlemoyer, and Veselin Stoyanov. 2019.
\newblock \href {http://arxiv.org/abs/1907.11692} {Roberta: A robustly
  optimized bert pretraining approach}.

\bibitem[{Maynez et~al.(2020)Maynez, Narayan, Bohnet, and
  McDonald}]{maynez-etal-2020-faithfulness}
Joshua Maynez, Shashi Narayan, Bernd Bohnet, and Ryan McDonald. 2020.
\newblock \href {https://doi.org/10.18653/v1/2020.acl-main.173} {On
  faithfulness and factuality in abstractive summarization}.
\newblock In \emph{Proceedings of the 58th Annual Meeting of the Association
  for Computational Linguistics}, pages 1906--1919, Online. Association for
  Computational Linguistics.

\bibitem[{McCoy et~al.(2019)McCoy, Pavlick, and Linzen}]{mccoy-etal-2019-right}
Tom McCoy, Ellie Pavlick, and Tal Linzen. 2019.
\newblock \href {https://doi.org/10.18653/v1/P19-1334} {Right for the wrong
  reasons: Diagnosing syntactic heuristics in natural language inference}.
\newblock In \emph{Proceedings of the 57th Annual Meeting of the Association
  for Computational Linguistics}, pages 3428--3448, Florence, Italy.
  Association for Computational Linguistics.

\bibitem[{Mishra et~al.(2021)Mishra, Patel, Vijayakumar, Li, Kapanipathi, and
  Talamadupula}]{mishra-etal-2021-looking}
Anshuman Mishra, Dhruvesh Patel, Aparna Vijayakumar, Xiang~Lorraine Li, Pavan
  Kapanipathi, and Kartik Talamadupula. 2021.
\newblock \href {https://doi.org/10.18653/v1/2021.naacl-main.104} {Looking
  beyond sentence-level natural language inference for question answering and
  text summarization}.
\newblock In \emph{Proceedings of the 2021 Conference of the North American
  Chapter of the Association for Computational Linguistics: Human Language
  Technologies}, pages 1322--1336, Online. Association for Computational
  Linguistics.

\bibitem[{Ni et~al.(2021)Ni, Ábrego, Constant, Ma, Hall, Cer, and
  Yang}]{ni2021sentencet5}
Jianmo Ni, Gustavo~Hernández Ábrego, Noah Constant, Ji~Ma, Keith~B. Hall,
  Daniel Cer, and Yinfei Yang. 2021.
\newblock \href {http://arxiv.org/abs/2108.08877} {Sentence-t5: Scalable
  sentence encoders from pre-trained text-to-text models}.

\bibitem[{Nie et~al.(2020)Nie, Williams, Dinan, Bansal, Weston, and
  Kiela}]{nie-etal-2020-adversarial}
Yixin Nie, Adina Williams, Emily Dinan, Mohit Bansal, Jason Weston, and Douwe
  Kiela. 2020.
\newblock \href {https://doi.org/10.18653/v1/2020.acl-main.441} {Adversarial
  {NLI}: A new benchmark for natural language understanding}.
\newblock In \emph{Proceedings of the 58th Annual Meeting of the Association
  for Computational Linguistics}, pages 4885--4901, Online. Association for
  Computational Linguistics.

\bibitem[{Pagnoni et~al.(2021)Pagnoni, Balachandran, and
  Tsvetkov}]{pagnoni-etal-2021-understanding}
Artidoro Pagnoni, Vidhisha Balachandran, and Yulia Tsvetkov. 2021.
\newblock \href {https://doi.org/10.18653/v1/2021.naacl-main.383}
  {Understanding factuality in abstractive summarization with {FRANK}: A
  benchmark for factuality metrics}.
\newblock In \emph{Proceedings of the 2021 Conference of the North American
  Chapter of the Association for Computational Linguistics: Human Language
  Technologies}, pages 4812--4829, Online. Association for Computational
  Linguistics.

\bibitem[{Poliak(2020)}]{poliak-2020-survey}
Adam Poliak. 2020.
\newblock \href {https://doi.org/10.18653/v1/2020.eval4nlp-1.10} {A survey on
  recognizing textual entailment as an {NLP} evaluation}.
\newblock In \emph{Proceedings of the First Workshop on Evaluation and
  Comparison of NLP Systems}, pages 92--109, Online. Association for
  Computational Linguistics.

\bibitem[{Poliak et~al.(2018)Poliak, Naradowsky, Haldar, Rudinger, and
  Van~Durme}]{poliak-etal-2018-hypothesis}
Adam Poliak, Jason Naradowsky, Aparajita Haldar, Rachel Rudinger, and Benjamin
  Van~Durme. 2018.
\newblock \href {https://doi.org/10.18653/v1/S18-2023} {Hypothesis only
  baselines in natural language inference}.
\newblock In \emph{Proceedings of the Seventh Joint Conference on Lexical and
  Computational Semantics}, pages 180--191, New Orleans, Louisiana. Association
  for Computational Linguistics.

\bibitem[{Raffel et~al.(2020)Raffel, Shazeer, Roberts, Lee, Narang, Matena,
  Zhou, Li, and Liu}]{t5_paper}
Colin Raffel, Noam Shazeer, Adam Roberts, Katherine Lee, Sharan Narang, Michael
  Matena, Yanqi Zhou, Wei Li, and Peter~J. Liu. 2020.
\newblock \href {http://jmlr.org/papers/v21/20-074.html} {Exploring the limits
  of transfer learning with a unified text-to-text transformer}.
\newblock \emph{Journal of Machine Learning Research}, 21(140):1--67.

\bibitem[{Roberts et~al.(2022)Roberts, Chung, Levskaya, Mishra, Bradbury,
  Andor, Narang, Lester, Gaffney, Mohiuddin, Hawthorne, Lewkowycz, Salcianu,
  van Zee, Austin, Goodman, Soares, Hu, Tsvyashchenko, Chowdhery, Bastings,
  Bulian, Garcia, Ni, Chen, Kenealy, Clark, Lee, Garrette, Lee-Thorp, Raffel,
  Shazeer, Ritter, Bosma, Passos, Maitin-Shepard, Fiedel, Omernick, Saeta,
  Sepassi, Spiridonov, Newlan, and Gesmundo}]{roberts2022scaling}
Adam Roberts, Hyung~Won Chung, Anselm Levskaya, Gaurav Mishra, James Bradbury,
  Daniel Andor, Sharan Narang, Brian Lester, Colin Gaffney, Afroz Mohiuddin,
  Curtis Hawthorne, Aitor Lewkowycz, Alex Salcianu, Marc van Zee, Jacob Austin,
  Sebastian Goodman, Livio~Baldini Soares, Haitang Hu, Sasha Tsvyashchenko,
  Aakanksha Chowdhery, Jasmijn Bastings, Jannis Bulian, Xavier Garcia, Jianmo
  Ni, Andrew Chen, Kathleen Kenealy, Jonathan~H. Clark, Stephan Lee, Dan
  Garrette, James Lee-Thorp, Colin Raffel, Noam Shazeer, Marvin Ritter, Maarten
  Bosma, Alexandre Passos, Jeremy Maitin-Shepard, Noah Fiedel, Mark Omernick,
  Brennan Saeta, Ryan Sepassi, Alexander Spiridonov, Joshua Newlan, and Andrea
  Gesmundo. 2022.
\newblock \href {http://arxiv.org/abs/2203.17189} {Scaling up models and data
  with $\texttt{t5x}$ and $\texttt{seqio}$}.

\bibitem[{Sainz et~al.(2022)Sainz, Gonzalez-Dios, de~Lacalle, Min, and
  Agirre}]{Sainz2022textual}
Oscar Sainz, Itziar Gonzalez-Dios, Oier~Lopez de~Lacalle, Bonan Min, and Eneko
  Agirre. 2022.
\newblock \href {https://doi.org/10.48550/ARXIV.2205.01376} {Textual entailment
  for event argument extraction: Zero- and few-shot with multi-source
  learning}.

\bibitem[{Schuster et~al.(2021)Schuster, Fisch, and
  Barzilay}]{schuster-etal-2021-get}
Tal Schuster, Adam Fisch, and Regina Barzilay. 2021.
\newblock \href {https://doi.org/10.18653/v1/2021.naacl-main.52} {Get your
  vitamin {C}! robust fact verification with contrastive evidence}.
\newblock In \emph{Proceedings of the 2021 Conference of the North American
  Chapter of the Association for Computational Linguistics: Human Language
  Technologies}, pages 624--643, Online. Association for Computational
  Linguistics.

\bibitem[{Schuster et~al.(2020)Schuster, Schuster, Shah, and
  Barzilay}]{schuster-etal-2020-limitations}
Tal Schuster, Roei Schuster, Darsh~J. Shah, and Regina Barzilay. 2020.
\newblock \href {https://doi.org/10.1162/coli_a_00380} {The limitations of
  stylometry for detecting machine-generated fake news}.
\newblock \emph{Computational Linguistics}, 46(2):499--510.

\bibitem[{Schuster et~al.(2019)Schuster, Shah, Yeo, Roberto Filizzola~Ortiz,
  Santus, and Barzilay}]{schuster-etal-2019-towards}
Tal Schuster, Darsh Shah, Yun Jie~Serene Yeo, Daniel Roberto Filizzola~Ortiz,
  Enrico Santus, and Regina Barzilay. 2019.
\newblock \href {https://doi.org/10.18653/v1/D19-1341} {Towards debiasing fact
  verification models}.
\newblock In \emph{Proceedings of the 2019 Conference on Empirical Methods in
  Natural Language Processing and the 9th International Joint Conference on
  Natural Language Processing (EMNLP-IJCNLP)}, pages 3419--3425, Hong Kong,
  China. Association for Computational Linguistics.

\bibitem[{Shah et~al.(2020)Shah, Schuster, and
  Barzilay}]{Shah_Schuster_Barzilay_2020}
Darsh Shah, Tal Schuster, and Regina Barzilay. 2020.
\newblock \href {https://doi.org/10.1609/aaai.v34i05.6406} {Automatic
  fact-guided sentence modification}.
\newblock \emph{Proceedings of the AAAI Conference on Artificial Intelligence},
  34(05):8791--8798.

\bibitem[{Storks et~al.(2019)Storks, Gao, and Chai}]{storks2019recent}
Shane Storks, Qiaozi Gao, and Joyce~Y. Chai. 2019.
\newblock \href {http://arxiv.org/abs/1904.01172} {Recent advances in natural
  language inference: A survey of benchmarks, resources, and approaches}.

\bibitem[{Tay et~al.(2021)Tay, Dehghani, Abnar, Shen, Bahri, Pham, Rao, Yang,
  Ruder, and Metzler}]{tay2021long}
Yi~Tay, Mostafa Dehghani, Samira Abnar, Yikang Shen, Dara Bahri, Philip Pham,
  Jinfeng Rao, Liu Yang, Sebastian Ruder, and Donald Metzler. 2021.
\newblock \href {https://openreview.net/forum?id=qVyeW-grC2k} {Long range arena
  : A benchmark for efficient transformers}.
\newblock In \emph{International Conference on Learning Representations}.

\bibitem[{Thorne et~al.(2018)Thorne, Vlachos, Christodoulopoulos, and
  Mittal}]{thorne-etal-2018-fever}
James Thorne, Andreas Vlachos, Christos Christodoulopoulos, and Arpit Mittal.
  2018.
\newblock \href {https://doi.org/10.18653/v1/N18-1074} {{FEVER}: a large-scale
  dataset for fact extraction and {VER}ification}.
\newblock In \emph{Proceedings of the 2018 Conference of the North {A}merican
  Chapter of the Association for Computational Linguistics: Human Language
  Technologies, Volume 1 (Long Papers)}, pages 809--819, New Orleans,
  Louisiana. Association for Computational Linguistics.

\bibitem[{Utama et~al.(2022)Utama, Bambrick, Moosavi, and
  Gurevych}]{utama-etal-2022-falsesum}
Prasetya Utama, Joshua Bambrick, Nafise Moosavi, and Iryna Gurevych. 2022.
\newblock \href {https://doi.org/10.18653/v1/2022.naacl-main.199} {Falsesum:
  Generating document-level {NLI} examples for recognizing factual
  inconsistency in summarization}.
\newblock In \emph{Proceedings of the 2022 Conference of the North American
  Chapter of the Association for Computational Linguistics: Human Language
  Technologies}, pages 2763--2776, Seattle, United States. Association for
  Computational Linguistics.

\bibitem[{Utama et~al.(2021)Utama, Moosavi, Sanh, and
  Gurevych}]{utama-etal-2021-avoiding}
Prasetya Utama, Nafise~Sadat Moosavi, Victor Sanh, and Iryna Gurevych. 2021.
\newblock \href {https://doi.org/10.18653/v1/2021.emnlp-main.713} {Avoiding
  inference heuristics in few-shot prompt-based finetuning}.
\newblock In \emph{Proceedings of the 2021 Conference on Empirical Methods in
  Natural Language Processing}, pages 9063--9074, Online and Punta Cana,
  Dominican Republic. Association for Computational Linguistics.

\bibitem[{Utama et~al.(2020)Utama, Moosavi, and
  Gurevych}]{utama-etal-2020-towards}
Prasetya~Ajie Utama, Nafise~Sadat Moosavi, and Iryna Gurevych. 2020.
\newblock \href {https://doi.org/10.18653/v1/2020.emnlp-main.613} {Towards
  debiasing {NLU} models from unknown biases}.
\newblock In \emph{Proceedings of the 2020 Conference on Empirical Methods in
  Natural Language Processing (EMNLP)}, pages 7597--7610, Online. Association
  for Computational Linguistics.

\bibitem[{Vaswani et~al.(2017)Vaswani, Shazeer, Parmar, Uszkoreit, Jones,
  Gomez, Kaiser, and Polosukhin}]{attention_all_you_need}
Ashish Vaswani, Noam Shazeer, Niki Parmar, Jakob Uszkoreit, Llion Jones,
  Aidan~N Gomez, \L~ukasz Kaiser, and Illia Polosukhin. 2017.
\newblock \href
  {https://proceedings.neurips.cc/paper/2017/file/3f5ee243547dee91fbd053c1c4a845aa-Paper.pdf}
  {Attention is all you need}.
\newblock In \emph{Advances in Neural Information Processing Systems},
  volume~30. Curran Associates, Inc.

\bibitem[{Vlachos and Riedel(2014)}]{vlachos-riedel-2014-fact}
Andreas Vlachos and Sebastian Riedel. 2014.
\newblock \href {https://doi.org/10.3115/v1/W14-2508} {Fact checking: Task
  definition and dataset construction}.
\newblock In \emph{Proceedings of the {ACL} 2014 Workshop on Language
  Technologies and Computational Social Science}, pages 18--22, Baltimore, MD,
  USA. Association for Computational Linguistics.

\bibitem[{Vrandecic(2020)}]{unified_wiki}
Denny Vrandecic. 2020.
\newblock \href {http://arxiv.org/abs/2004.04733} {Architecture for a
  multilingual wikipedia}.
\newblock \emph{CoRR}, abs/2004.04733.

\bibitem[{Vu et~al.(2021)Vu, Luong, Le, Simon, and Iyyer}]{vu-etal-2021-strata}
Tu~Vu, Minh-Thang Luong, Quoc Le, Grady Simon, and Mohit Iyyer. 2021.
\newblock \href {https://doi.org/10.18653/v1/2021.emnlp-main.462} {{ST}ra{TA}:
  Self-training with task augmentation for better few-shot learning}.
\newblock In \emph{Proceedings of the 2021 Conference on Empirical Methods in
  Natural Language Processing}, pages 5715--5731, Online and Punta Cana,
  Dominican Republic. Association for Computational Linguistics.

\bibitem[{Williams et~al.(2018)Williams, Nangia, and
  Bowman}]{williams-etal-2018-broad}
Adina Williams, Nikita Nangia, and Samuel Bowman. 2018.
\newblock \href {https://doi.org/10.18653/v1/N18-1101} {A broad-coverage
  challenge corpus for sentence understanding through inference}.
\newblock In \emph{Proceedings of the 2018 Conference of the North {A}merican
  Chapter of the Association for Computational Linguistics: Human Language
  Technologies, Volume 1 (Long Papers)}, pages 1112--1122, New Orleans,
  Louisiana. Association for Computational Linguistics.

\bibitem[{Wu et~al.(2022)Wu, Gardner, Stenetorp, and Dasigi}]{Yuxiang2022}
Yuxiang Wu, Matt Gardner, Pontus Stenetorp, and Pradeep Dasigi. 2022.
\newblock \href {https://doi.org/10.48550/ARXIV.2203.12942} {Generating data to
  mitigate spurious correlations in natural language inference datasets}.

\bibitem[{Yin et~al.(2021)Yin, Radev, and Xiong}]{yin-etal-2021-docnli}
Wenpeng Yin, Dragomir Radev, and Caiming Xiong. 2021.
\newblock \href {https://doi.org/10.18653/v1/2021.findings-acl.435}
  {{D}oc{NLI}: A large-scale dataset for document-level natural language
  inference}.
\newblock In \emph{Findings of the Association for Computational Linguistics:
  ACL-IJCNLP 2021}, pages 4913--4922, Online. Association for Computational
  Linguistics.

\bibitem[{Zeng et~al.(2021)Zeng, Abumansour, and Zubiaga}]{zeng2021automated}
Xia Zeng, Amani~S Abumansour, and Arkaitz Zubiaga. 2021.
\newblock Automated fact-checking: A survey.
\newblock \emph{Language and Linguistics Compass}, 15(10):e12438.

\end{thebibliography}
